%% file: main.tex
\documentclass[10pt,journal,compsoc]{IEEEtran}
\ifCLASSOPTIONcompsoc
  \usepackage[nocompress]{cite}
\else
  \usepackage{cite}
\fi
\ifCLASSINFOpdf
\else
\fi

\usepackage{multirow}
\usepackage{booktabs,enumitem,url}
\usepackage{amssymb,amsmath,amsthm}
\usepackage{makecell}
\usepackage{graphicx}
\usepackage{subfigure}
\usepackage{pifont}
\input{math_commands}

\usepackage{hyperref}
\usepackage{cleveref}
\usepackage{adjustbox}

\newcommand{\cmark}{\ding{51}}

\newcommand{\ie}{\em{i.e.}}
\newcommand{\eg}{\em{e.g.}}

\usepackage{xcolor}

\newcommand{\firsttask}{\textit{de novo} 2D molecule generation}
\newcommand{\secondtask}{\textit{de novo} 3D conformation generation}
\newcommand{\thirdtask}{\textit{de novo} binding-based 3D molecule generation}
\newcommand{\fourthtask}{\textit{de novo} binding-pose conformation generation}
\newcommand{\fifthtask}{2D molecule optimization}
\newcommand{\sixthtask}{\textit{de novo} 2D molecule optimization}
\newcommand{\seventhtask}{\textit{de novo} 3D molecule generation}
\newcommand{\eighthtask}{3D molecule optimization}

\begin{document}
%
\title{MolGenSurvey: \\
A Systematic Survey in Machine Learning Models for Molecule Design}
%
%
%
%

\author{Yuanqi Du$^{*}$,
        Tianfan Fu$^{*}$,
        Jimeng Sun,
        Shengchao Liu
\IEEEcompsocitemizethanks{
\IEEEcompsocthanksitem Y. Du and T. Fu contribute equally to this work.
\IEEEcompsocthanksitem Y. Du is with George Mason University, Fairfax, VA, US 22030.
\IEEEcompsocthanksitem T. Fu is with Georgia Institute of Technology, Atlanta, GA, US 30332.
\IEEEcompsocthanksitem J. Sun is with University of Illinois at Urbana-Champaign, Champaign, IL, US 61820.
\IEEEcompsocthanksitem S. Liu is with Mila - Université de Montréal, Montréal, QC, Canada H2S 3H1.
\IEEEcompsocthanksitem Correspondence to S. Liu.
}
\thanks{Preprint.}}

%
%

\markboth{Journal of \LaTeX\ Class Files,~Vol.~14, No.~8, August~2015}%
{Shell \MakeLowercase{\textit{et al.}}: Bare Demo of IEEEtran.cls for Computer Society Journals}
%



\IEEEtitleabstractindextext{%
\begin{abstract}
Molecule design is a fundamental problem in molecular science and has critical applications in a variety of areas, such as drug discovery, material science, etc. However, due to the large searching space, it is impossible for human experts to enumerate and test all molecules in wet-lab experiments.
Recently, with the rapid development of machine learning methods, especially generative methods, molecule design has achieved great progress by leveraging machine learning models to generate candidate molecules. 
In this paper, we systematically review the most relevant work in machine learning models for molecule design.
We start with a brief review of the mainstream molecule featurization and representation methods (including 1D string, 2D graph, and 3D geometry) and general generative methods (deep generative and combinatorial optimization methods).
Then we summarize all the existing molecule design problems into several venues according to the problem setup, including input, output types and goals.
Finally, we conclude with the open challenges and point out future opportunities of machine learning models for molecule design in real-world applications.
\end{abstract}

}

\maketitle

\IEEEdisplaynontitleabstractindextext

%
\IEEEpeerreviewmaketitle

\input{01_intro}

\input{02_preliminaries}

\input{03_generative_models}

\input{04_tasks}
\input{05_evaluations}

\input{06_future_directions}

\input{07_conclusion}


\bibliographystyle{IEEEtran}
\bibliography{main}







\end{document}

%% file: math_commands.tex
\usepackage{amsmath,amsfonts,bm}

\def\eqref#1{equation~\ref{#1}}

\def\1{\bm{1}}

\def\vx{{\bm{x}}}

\def\vz{{\bm{z}}}

\DeclareMathAlphabet{\mathsfit}{\encodingdefault}{\sfdefault}{m}{sl}
\SetMathAlphabet{\mathsfit}{bold}{\encodingdefault}{\sfdefault}{bx}{n}



%% file: 01_intro.tex
\section{Introduction} 
\label{sec:intro}

Designing new molecules has been a long-standing challenge in molecular science and has critical applications in many domains such as drug discovery and material design~\cite{drews2000drug}. 
Traditional methods, e.g., high-throughput screening (HTS) technologies, search over existing molecule databases but rely heavily on enumeration, thus are very time-consuming and laborious~\cite{dimasi2016innovation,wouters2020estimated}. 
On the other hand, the number of the drug-like molecules is large as estimated to be $10^{60}$~\cite{polishchuk2013estimation} and it is computationally prohibitive to enumerate all the possible molecules. 
To address the issue, machine learning approaches have been proposed to narrow down the searching space of molecules and explore the chemical space intelligently.  

Among different types of machine learning approaches, generative methods offer a promising direction for the automated design of molecules with desired properties such as synthesis accessibility, binding affinity, etc. A wide range of generative methods has been proposed for molecule design, mainly in two types, (1) deep generative models and (2) combinatorial optimization methods. 
Deep generative models (DGMs) usually map molecules into a continuous latent space while combinatorial optimization methods directly search over the discrete chemical space. 

Another key component of molecule design in machine learning is how the molecules are represented which includes 1D string, 2D graph, 3D geometry, etc. Early attempts formulate the molecule design as a 1D string generation problem and represent molecules as simplified molecular-input line-entry system (SMILES) strings. 
However, the validity of the generated SMILES string is not guaranteed~\cite{gomez2018automatic}. 
To address the issues, other string featurization methods are explored, including an alternative string representation, self-referencing embedded strings (SELFIES)~\cite{nigam2019augmenting}, which enforces string validity by design. With the emergence of graph representation learning, 2D graph representation has been introduced for molecule design and soon become a research hot spot as 2D graph is expected to carry richer structure information than 1D string.

Despite the success achieved in 1D and 2D molecule generation, molecules are naturally 3D structures and 3D conformations of molecules are crucial to many real-world applications such as structured-based drug design. However, representing a molecule by its 3D geometry is a nontrivial problem as it has to respect symmetry in 3D space such as translation- and rotation-equivariance.
With the recent advances on geometric deep learning~\cite{bronstein2017geometric}, efforts have been put to study representing molecules by 3D geometry and the wave has led to advances in molecule design in 3D space.

Nevertheless, one of the biggest challenges that stifles the progress of machine learning for molecule design is the fast-growing and not well-organized literature. Specifically, many different types of featurization and generative methods have been proposed for different types of molecule design tasks while the boundaries among them are not clearly specified.
Although many recent efforts have been put to review the literature of machine learning models for molecule design, most of them only focus on part of the literature in terms of generative methods and tasks~\cite{sanchez2018inverse,xue2019advances,elton2019deep,vanhaelen2020advent,alshehri2020deep,jimenez2020drug,axelrod2022learning}.
To the best of our knowledge, we are the first to provide a comprehensive review on machine learning approaches (both deep generative and combinatorial optimization methods) for 1D/2D/3D molecule design under a variety of task settings. Specifically, we present a mind map that drives the key taxonomies in molecule design tasks and unifies the task settings via the input/output types, and goals. The main contributions are summarized as follows:

\begin{itemize}
\item \textbf{Review of Methods.} Section~\ref{sec:representation} and~\ref{sec:method} gather and categorize the mainstream molecule representation and general generation approaches, respectively.
\item \textbf{Review of Tasks.} Section~\ref{sec:task} elaborates the existing molecule design tasks and unifies them into a highly structured mindmap/table.
\item \textbf{Review of Experiments.} Section~\ref{sec:evaluation} specifies the empirical setups of molecule design, including evaluation metrics, available datasets, oracles (usually molecular properties), etc.  
\item \textbf{Challenges and Opportunities.} Section~\ref{sec:future} identifies open challenges and opportunities along the research direction for molecule design, paving the way for the future of molecule design with machine learning models.
\end{itemize}

%% file: 02_preliminaries.tex
\section{Molecule Structure and Representation Learning} 
\label{sec:representation}

\begin{figure}[htbp]
\centering
\begin{subfigure}[2D Molecular Graph]
{\includegraphics[width=.45\linewidth]{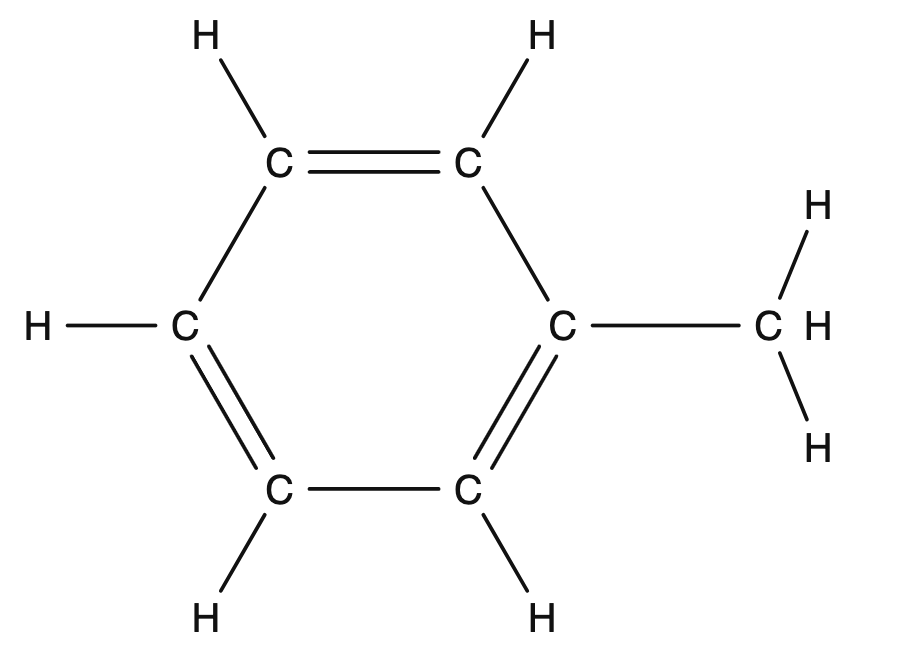} \label{fig:molecule_feature_example_2D}}
\end{subfigure}
\begin{subfigure}[3D molecular Geometry]
{\includegraphics[width=.45\linewidth]{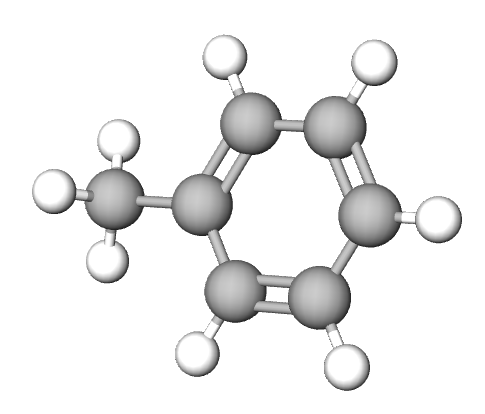} \label{fig:molecule_feature_example_3D}}
\end{subfigure}
\caption[Caption for LOF]{
    Four data structures on an example molecule.
    The 1D SMILES string is Cc1ccccc1, and the 1D SELFIES string is [C][C][=C][C][=C][C][=C][Ring1][=Branch1].
    \Cref{fig:molecule_feature_example_2D,fig:molecule_feature_example_3D} are the 2D and 3D molecular Geometry respectively.
\footnotemark}
\label{fig:molecule_feature_examples}
\end{figure}

In this section, we introduce the mainstream featurization and representation learning methods for molecules. The three prevalent features adopted in molecule generation tasks are 1D string description, 2D molecular graph, and 3D molecular geometry. One example of the three features on a specific molecule (IUPAC: methylbenzene) is given in~\Cref{fig:molecule_feature_examples}.

\footnotetext{Credits to 2D and 3D figures are from \url{https://molview.org/?q=Cc1ccccc1}.}

\subsection{1D String Description}
The 1D string description is to extract molecules into a 1D string following certain domain-specific rules. The mainstream 1D string descriptions used for molecule generation include simplified molecular-input line-entry system (SMILES)~\cite{weininger1988smiles} and self-referencing embedded strings (SELFIES)~\cite{krenn2020self}. SMILES is well-known for its simplicity in denoting molecules as strings by following rules like adjacent atoms are assumed to be connected by a single or aromatic bond and branches are specified in parentheses, etc. However, one major limitation is that a significant fraction of SMILES strings are invalid. To avoid this, SELFIES augments the rules on handling branches and rings and thus it can achieve 100\% validity.

We want to acknowledge that there exist other variants of 1D string descriptions, including International Chemical Identifier (InChI)~\cite{heller2015inchi}, SMILES Arbitrary Target Specification (SMARTS)~\cite{smarts}; yet, SMILES and SELFIES are the two most commonly-used for molecule generation tasks.

\textbf{Representation methods.} For molecular string representation, existing work considers advanced representation methods from other domains, including convolutional neural network (CNN)~\cite{hirohara2018convolutional}, recurrent neural network (RNN)~\cite{bjerrum2017smiles,liu2018practical,huang2020deeppurpose}, and Transformer~\cite{honda2019smiles,wang2019smiles,chithrananda2020chemberta}.

\subsection{2D Molecular Graph}
A large fraction of molecules can be naturally treated as 2D graphs: the atoms and bonds are nodes and edges in the graph, respectively. One example is given in~\Cref{fig:molecule_feature_example_2D}. Such featurization method has brought wide explorations in the structured data representation and related applications.

\textbf{Representation methods.}
Graph neural network (GNN) is leveraged for 2D molecular graph representation. The high-level idea can be summarized into the \textit{message-passing neural network (MPNN)} framework~\cite{gilmer2017neural}: it updates each node's representation by aggregating information from its neighborhood; by repeating this for $K$ steps, then each node's final representations can reveal the information in the $K$-hop neighborhood. Generally, graph application is a broad field where certain GNNs~\cite{kipf2016semi,hamilton2017inductive,xu2018powerful} are originally proposed to solve other structured data, {\eg}, social network and citation network. Specifically for the molecular graph representation, NEF~\cite{duvenaud2015convolutional} is the first to apply GNN for property prediction, and more recent works have considered different aspects for improving the expressiveness of GNNs. Weave~\cite{kearnes2016molecular} explicitly models both the node- and edge-level representation. D-MPNN~\cite{yang2019analyzing} and CMPNN~\cite{song2020communicative} explicitly add the direction information in message passing. PNA~\cite{corso2020principal} considers various aggregation methods during message passing. N-Gram Graph~\cite{liu2018n}, on the other hand, proposes the walk-based GNN that can alleviate the over-squashing issue~\cite{alon2020bottleneck} in the MPNN-based GNN. Another big track is motivated by the recent success of the attention module in the NLP and vision domains~\cite{vaswani2017attention,devlin2018bert,dosovitskiy2020image}. To adapt it to molecular graphs, recent works utilize the attention to better capture the correlation at node-, subgraph-, and graph-level, including GAT~\cite{velivckovic2017graph}, AttentiveFP~\cite{xiong2019pushing}, GTransformer~\cite{rong2020self}, AWARE~\cite{demirel2021analysis}, and Graphormer~\cite{ying2021transformers}.

\subsection{3D molecular Geometry}
Essentially, most molecule are not static, but instead in continual motion on a potential energy surface~\cite{axelrod2020geom}, characterized by different conformations, and the 3D structures at the local minima on this surface are named \textit{conformer}\footnote{A more rigorous way of defining conformer is in~\cite{moss1996basic}: a conformer is an isomer of a molecule that differs from another isomer by the rotation of a single bond in the molecule.}. More concretely, we can view it as a sets of nodes in the 3D Euclidean space, as displayed in~\Cref{fig:molecule_feature_example_3D}.

\textbf{Representation methods.}
Recent work on representing the 3D molecular Geometry assumes that if the learned representation is E(3)- or SE(3)-equivariant to a set of transformations, then such representation can reveal better expressiveness. This research line includes SchNet~\cite{schutt2018schnet}, OrbNet~\cite{qiao2020orbnet}, DimeNet~\cite{klicpera2020directional,klicpera2020fast}, TFN~\cite{fuchs2020se}, SE(3)-Transformer~\cite{fuchs2020se}, EGNN~\cite{satorras2021n}, SpinConv~\cite{shuaibi2021rotation}, SphereNet~\cite{liu2021spherical}, EVFN~\cite{du2021equivariant}, UNiTE~\cite{qiao2021unite}, SEGNN~\cite{brandstetter2021geometric}, PaiNN~\cite{schutt2021equivariant}, and GemNet~\cite{klicpera_gemnet_2021}.

\subsection{Other Molecular Featurization Methods}
There also exist other molecule featurization and representation methods, yet they are not fit or not being widely explored for molecule generation tasks. Here we may as well briefly list them for comprehensiveness.

\textbf{Descriptors} It is mainly obtained by hashing the sub-components of the molecule into a fixed-length vector. The representation methods on fingerprints include random forest, XGBoost, multi-layer perceptron, etc. Recent work~\cite{ramsundar2015massively,liu2018practical,alnammi2021evaluating,jiang2021could} have empirically verified the effectiveness of descriptors. However, due to the irreversible hashing operation, it cannot be directly utilized for generation tasks.

\textbf{3D density map} This represents molecules as the 3D volume occupied by each of the atoms~\cite{orlando2022pyuul}. The volume for each atom is usually a function of the atomic radius which makes it a sphere-like solid object. The density map is often stored in a 3D grid. 3D CNN is a typical neural network utilized to encode the 3D density map~\cite{francoeur2020three}.

\textbf{3D surface mesh} This represents molecules by their surfaces with 3D meshes (e.g., polygon mesh). The mesh consists of a collection of nodes, edges, and faces that define the shape of the molecular surface. The 3D surface mesh is usually encoded by a geodesic Convolutional Neural Network~\cite{masci2015geodesic} or 3D Graph Neural Network~\cite{schutt2018schnet}.

\textbf{Chemical Images} This extracts the molecules into 2D grid images, as known as the chemical images. Then standard CNNs~\cite{goh2017chemception,meyer2019learning} are applied on it for representation. Yet, this hasn't been explored for molecule generation tasks yet.

\subsection{Summary}
Input data featurization and the learned representation function are two key components for molecule generation. A lot of follow-up works are built on this, including multi-task learning~\cite{ramsundar2015massively,liu2018exploration,liu2019loss,liu2021multitask,liu2020structured}, self-supervised learning~\cite{hu2019strategies,sun2019infograph,liu2021pre,fu2021spear,xia2022survey}, and few-shot learning~\cite{altae2017low,baskin2019one,nguyen2020meta}. In the following sections, we will see how different feature and representation methods may fit into the generative methods (\Cref{sec:generative_models}) and molecule generation tasks (\Cref{sec:task}).

%% file: 03_generative_models.tex
\section{Generative Methods} 
\label{sec:generative_models} 
\label{sec:method}

There are two main venues for generative methods. The first one is deep generative models (DGMs), which aim at modeling the data distribution using deep learning models. The second one is combinatorial optimization methods (COMs) that better utilize the heuristic strategies for the target goal. Specifically for the structured data, the main difference between these two venues is that DGMs utilize the \textbf{continuous latent representation}, while the COMs are directly searching in the \textbf{structured data space}. More details are discussed below, and key notations are listed in~\Cref{tab:notation}.

\begin{table}[h]
\centering
\caption{Key notations for generative methods.}
\label{tab:notation}
\begin{adjustbox}{max width=\textwidth}
\begin{tabular}{l l}
\toprule
Notation & Description \\
\midrule
$\vx$ & data point (2D or 3D molecule) \\
$\vz$ & latent representation\\
\midrule
For AR:\\
$d$ & \# subcomponents of data point $\vx$\\
$\bar x_0, \bar x_1, ..., \bar x_{d-1}$ & each subcomponent of data point $\vx$\\
\midrule
For NF:\\
$K$ & \# latent representation\\
$\vz_0, \vz_1, ..., \vz_K$ & list of $K$ latent representations\\
\midrule
For diffusion model:\\
$T$ & \# steps\\
$\vx_1, \vx_2, ..., \vx_T$ & $T$ latent samples\\
\bottomrule
\end{tabular}
\end{adjustbox}
\end{table}

\subsection{Deep Generative Models (DGMs)}
\label{sec:dgm}
\subsubsection{Autoregressive Models (ARs)}
For each data point $\vx$, especially for structured data like a graph, we first assume it has $d$ subcomponents, and such subcomponents can have underlying dependence. For example, the subcomponents can be the pixels in images or nodes and bonds in molecular graphs. The autoregressive models (ARs) learn the joint distribution of $\vx$ by factorizing it as the product of $d$ subcomponent likelihoods as below:
\begin{equation} \small{ \label{eq:autoregressive}
p(\vx) = \prod_{i=1}^d p(\bar x_i|\bar x_1, \bar x_2, ..., \bar x_{i-1}).
} \end{equation}
Here ARs define an ordering of variables $\bar x_i$ (either heuristically or with domain knowledge), and model the joint distribution in an \textbf{autoregressive} or \textbf{sequential} manner: predicting the next subcomponent based on the previous subcomponents. ARs have been widely-used in generation tasks, including recurrent neural network (RNN)~\cite{rumelhart1986learning,hochreiter1996lstm}, PixelCNN~\cite{van2016pixel}, WaveNet~\cite{oord2016wavenet}, Transformer~\cite{vaswani2017attention} and BERT~\cite{devlin2018bert}.

\subsubsection{Variational Autoencoders (VAEs)} \label{sec:vae}
Variational autoencoders (VAEs)~\cite{kingma2013auto} estimate a lower bound (a.k.a. evidence lower bound (ELBO)) of the likelihood $p(\vx)$. VAEs introduce a proposal distribution $q(\vz|\vx)$ and the goal is to maximize the ELBO:
\begin{equation} \small{ \label{eq:variation_autoencoder}
\begin{aligned}
\log p(\vx)
& = \log \int_\vz p(\vz) p(\vx|\vz) d\vz\\
& \ge \mathbb{E}_{q(\vz|\vx)} \big[ \log p(\vx|\vz) \big] - D_{KL}(q(\vz|\vx) || p(\vz))\\
& \triangleq \text{ELBO}.
\end{aligned}
} \end{equation}
The first term on ELBO is also known as the reconstruction term. Classic VAEs~\cite{kingma2013auto} take $\vx$ as either continuous or binary, where we often model this conditional term as Gaussian or Bernoulli distribution. This will lead to mean-squared error (MSE) or binary cross-entropy (BCE) for reconstruction.

The second term of~\Cref{eq:variation_autoencoder} is the KL-divergence term measuring the information loss when using $q(\vz|\vx)$ to approximate $p(\vz)$. In VAEs, the standard Gaussian is often used as the prior, {\ie}, $p(\vz) \sim \mathcal{N}(0, I)$. Such standard Gaussian implies that all the learned latent dimensions are expected to be diverse or \textbf{disentangled}. Following work, {\eg}, $\beta$-VAE~\cite{higgins2016beta}, penalizes this KL-divergence regularizer term to encourage the more disentangled latent representation, such that~\cite{bengio2013representation} each single unit in the latent representation is sensitive to only certain factors.

\textbf{Reparameterization trick}. VAEs express $\vz$ as a deterministic variable $\vz = g(\epsilon, \vx)$, where $\epsilon$ is an auxiliary variable with independent marginal $p(\epsilon)$, and $g(\cdot)$ is the reconstruction function. Then $\vz$ can be reparameterized as:
\begin{equation} \small{ \label{eq:VAE_reparameterization}
\begin{aligned}
\vz = g_\phi(\vx, \epsilon) = \mu_{\vx} + \sigma_{\vx} \odot \epsilon,
\end{aligned}
} \end{equation}
where $\odot$ is the element-wise product. This reparameterization trick allows us to calculate the gradient of the reconstruction function with lower variance.

\subsubsection{Normalizing Flows (NFs)} \label{sec:flow}
The normalizing flows (NFs) aim at modeling the data distribution $p(\vx)$ directly with an invertible and deterministic mapping. First NFs suppose $\vx$ and $\vz$ are two random variables with the same dimension, $\vx \sim p(\vx)$ and $\vz \sim p(\vz)$. There exists an invertible mapping between the two variables, {\ie}, $\vx=f(\vz)$ and $\vz=f^{-1}(\vx)$. Then according to the \textbf{change of variable theorem}, we can have
\begin{equation} \small{\label{eq:change_of_variable_theorem}
p(\vx) = p(\vz) \Big| \text{det} \Big( \frac{d f^{-1}(\vx)}{d \vx} \Big) \Big|,
} \end{equation}
where $\frac{d f^{-1}(\vx)}{d \vx}$ is the Jacobian matrix and $\text{det}(\cdot)$ is the matrix determinant.

Based on ~\Cref{eq:change_of_variable_theorem}, an NFs map a simple prior distribution $p_0(\vz_0)$ to a complex data distribution $p(\vx)$ by applying a sequence of invertible transformation functions. In other words, the NFs flow through a chain of $K$ transformations $\vz_0 \to \vz_1 \to ... \to \vz_K = \vx$. Thus, the data distribution can be computed as
\begin{equation} \small{ \label{eq:normalizing_flow_model}
\begin{aligned}
\log p(\vx)
& = \log p_{K} (\vz_K)\\
& = \log p_{K-1} (\vz_{K-1}) - \log \Big| \text{det} \Big( \frac{d f_K(\vz_{K-1})}{d \vz_{K-1}} \Big) \Big|\\
& = ...\\
& = \log p_0 (\vz_0) - \sum_{i=1}^K \log \Big| \text{det} \Big( \frac{d f_{i}(\vz_{i-1})}{d \vz_{i-1}} \Big) \Big|.
\end{aligned}
} \end{equation}
Existing works, {\eg}, RealNVP~\cite{dinh2016density}, NICE~\cite{dinh2014nice}, and Glow~\cite{kingma2018glow}, judiciously design invertible transformations $f_i$, where the Jacobian determinant is simple to compute.

\subsubsection{Generative Adversarial Networks (GANs)}
The Generative Adversarial Networks (GANs)~\cite{goodfellow2014generative} is a likelihood-free deep generative model. GANs do not optimize the likelihood directly; instead, they aim at detecting the real data from fake data, {\ie}, they transform the generation problem into a binary classification problem.

There are two main components in GANs: one generator ($G$) and one discriminator ($D$). The generator generates data points to fool the discriminator, while the discriminator tries to distinguish the generated data from the actual data. The objective of GANs is:
\begin{equation} \small{ \label{eq:generative_adversarial_net}
\begin{aligned}
& \min_{G} \max_{D} \mathcal{L}(D, G) \\
= & \, \mathbb{E}_{\vx \sim p_{data} } [\log D(\vx)] + \mathbb{E}_{\vz \sim p(\vz)} [\log (1-D(G(\vz))].
\end{aligned}
} \end{equation}
The training of GANs risks the modal and model collapse issue. To alleviate this, f-GAN~\cite{nowozin2016f} adopts the f-divergence for measuring distribution distance, where the f-divergence is a more general metric that covers other widely-used distances, including KL divergence and Jenson-Shannon divergence. An alternative is Wasserstein-GAN (WGAN)~\cite{arjovsky2017wasserstein} which utilizes the Wasserstein distance for measuring the distance between distributions. Follow-up works, including StyleGAN~\cite{karras2019style} and StyleGAN2~\cite{karras2020analyzing}, introduce certain inductive biases in the architecture design and have been widely adopted in the community.

\subsubsection{Diffusion models}
Diffusion models~\cite{sohl2015deep,ho2020denoising} are inspired by non-equilibrium thermodynamics and can be split into the forward and backward diffusion processes. During the forward diffusion process, the diffusion models will gradually add noise to the data and the last-step data will follow an isotropic Gaussian. The reverse diffusion process will revert this process and construct the data from noise distribution.

More rigorously, we can define the forward process as from the actual data $\vx_0 \sim p(\vx)$ to the random noise $\vx_T$ with $T$ diffusion steps. Let us first assume that for the forward process, the Gaussian distribution is
\begin{equation*} \small{
\begin{aligned}
q(\vx_t | \vx_{t-1})
& = \mathcal{N}(\vx_{t}; \sqrt{1-\beta_t} \vx_{t-1}, \beta_t I),
\end{aligned}
} \end{equation*}
where $\beta_t \in (0, 1)$. Then the corresponding backward process is 
\begin{equation*} \small{ \label{eq:diffusion_backward}
\small{
\begin{aligned}
p_\theta(\vx_{t-1}|\vx_t) 
& = \mathcal{N}(\vx_{t-1}; \mu_\theta(\vx_t, t), \Sigma_\theta(\vx_t,t))\\
& = \mathcal{N}(\vx_{t-1}; \frac{1}{\sqrt{\alpha_t}} \big( \vx_t - \frac{\beta_t}{\sqrt{1 - \bar \alpha_t}} \epsilon \big), \frac{1-\bar \alpha_{t-1}}{1 - \bar \alpha_t} \beta_t),
\end{aligned}
}
} \end{equation*}
where $\epsilon \sim \mathcal{N}(0,I)$ follows the standard Gaussian, $\alpha_t = 1-\beta_t$, and $\bar \alpha_t = \prod_{i=1}^t \alpha_i$.
The objective is to estimate the variational lower bound (VLB) of the likelihood:
\begin{equation*} \small{
\begin{aligned}
\log p(\vx) \ge - \mathbb{E}_{q(x_{1:T}|x_0)} [\log \frac{ q(x_{1:T}|x_0) }{p_\theta(x_{0:T})}] = - \mathcal{L}_{\text{VLB}}.
\end{aligned}
} \end{equation*}
The VLB can be rewritten as:
\begin{equation*} \small{
\begin{aligned}
& \mathcal{L}_{\text{VLB}} 
 = \underbrace{KL[q(x_T|x_0) || p_\theta(x_T)]}_{\mathcal{L}_T} \\
& ~~~ + \sum_{t=2}^T \underbrace{KL[q(x_{t-1}|x_t,x_0) || p_\theta(x_{t-1}|x_t)]}_{\mathcal{L}_{t-1}} \underbrace{- \mathbb{E}_q [\log p_\theta(x_0|x_1)]}_{\mathcal{L}_0}.
\end{aligned}
} \end{equation*}
Here $\mathcal{L}_T$ is a constant and can be ignored, and existing methods~\cite{ho2020denoising,weng2021diffusion} have been using a separate model for estimating $\mathcal{L}_0$. For $\{\mathcal{L}_{t-1}\}_{t=2}^T$, they adopt a neural network to approximate the conditionals during the reverse process, and reduce to the following objective function:
\begin{equation*} \small{
\mathcal{L}_t = \mathbb{E}_{\vx_0,z} \Big[\| \epsilon_t - \epsilon_\theta(\sqrt{\bar \alpha_t}\vx_0 + \sqrt{1-\bar \alpha_t}\epsilon_t, t) \|^2 \Big].
} \end{equation*}
Please also notice that the final objective function is equivalent to a family of score matching methods~\cite{song2019generative,song2020score}, as will be introduced below.

\subsubsection{Energy-Based Models (EBMs)}
Energy-based models (EBMs) have a very flexible formulation, and its general form is:
\begin{equation} \small{\label{eq:energy_based_model}
p(\vx) = \frac{\exp(-E_{\theta}(\vx))}{A} = \frac{\exp(-E_{\theta}(\vx))}{\int_{\vx} \exp(-E_{\theta}(\vx)) d\vx},
} \end{equation}
where $E_{\theta}(\cdot)$ is the energy function, $\theta$ is the set of learnable parameters, and $A$ is the partition function. In EBMs, the bottleneck is the calculation of partition function $A$: it is computationally intractable due to the high cardinality of $\mathcal{X}$. Various methods have been proposed to handle this task, and the most widely-used ones include contrastive divergence~\cite{hinton2002training}, noise-contrastive estimation (NCE)~\cite{gutmann2010noise}, score matching (SM)~\cite{song2019generative}. Each method has been widely explored, and for more details, we refer to this comprehensive summary~\cite{song2021train}.

\subsection{Combinatorial Optimization Methods (COMs)} 
\label{sec:searching}

Deep generative models (DGMs) usually require a large amount of data to learn, which may be infeasible in real scenarios. 
In contrast, combinatorial optimization methods (COMs) have less requirement on the amount of training data, while the trade-off is that algorithm designers need to call the oracles during the exploration to the chemical space~\cite{zhou2019optimization,fu2021differentiable,gao2020synthesizability,gao2021amortized,brown2019guacamol}.
We mainly discuss five categories in COMs, (1) reinforcement learning (RL); (2) genetic algorithm (GA); (3) Bayesian Optimization (BO); (4) Monte Carlo Tree Search (MCTS), and (5) MCMC sampling. 
However, to conduct a thorough exploration of the chemical space, we need intensive oracle calls in combinatorial-based methods. 
For example, when we want to optimize molecules' Quantitative Estimate of Drug-likeness (QED) score during the generative process, we need to evaluate the QED score of the generated molecules to guide the next step actions ({\ie}, call QED oracle).

\subsubsection{Reinforcement Learning (RL)}
Reinforcement learning (RL) is usually formulated as a Markov Decision Process (MDP). 
RL has several basic concepts: state, action, transition, and reward. The state is denoted as $S$, action set is denoted as $A$. At the $t$-th step, the transition probability from state $s$ to state $s'$ under action $a \in A$ is $P_{\theta}(s, s') = P(s_{t+1} = s' | s_t = s, a_t = a)$, where $\theta$ is usually parameterized by neural networks. $R_a(s,s')$ is the reward after transition from $s$ to $s'$ with action $a$. 
The goal of RL is to learn an optimal policy that maximizes the ``reward'', 
\begin{equation} \small{
\underset{\theta}{\arg\max} \sum_{t=1}^{T} R_{a_t}(s_t, s_{t+1}). 
} \end{equation}
For molecule generation tasks, a state is usually defined as a partially generated molecule graph, action can be adding a substructure (either an atom or a bond or a ring) at a position, and reward can be the property of the partially generated molecule. 
Specifically, \cite{You2018-xh,jin2020multi,fu2021moler,bengio2021gflownet} leverage policy gradient and domain-specific reward function to optimize the graph policy neural network; 
\cite{zhou2019optimization} leverages deep Q-network to conduct value learning, where a deep Q-network is trained to predict the future reward and during each step, the step with maximal expected reward is selected.  

\subsubsection{Genetic Algorithm (GA)} 

Genetic Algorithm (GA) starts from a \textit{mating pool} (set of strings/graphs that serve as warm start) and requires \textit{mutation} and \textit{crossover} between parents (interaction between intermediates in the last iteration) to obtain the next generation, where the population in each iteration is called a \textit{generation}; 
\textit{parent} is the intermediate from the previous generation; 
\textit{crossover}, also called recombination, combines the genetic information of two parents to generate new offspring, offsprings are generated by cloning an existing solution; 
\textit{mutation} involves an arbitrary bit in a genetic sequence that will be flipped from its original state. A common method of the mutation operator involves generating a random variable for each bit in a sequence. 
Specifically, at the $t$-th generation (iteration), given a population of molecules, denoted as $\mathcal{P}^{(t-1)}$, we generate a bunch of offsprings (the set is denoted as $\mathcal{S}^{(t)}$) and select the most promising $K$ ($K$ molecules with highest objectives) as the population for the next generation (denoted as $\mathcal{P}^{(t)}$). It is defined as
\begin{equation} \small{
\label{eqn:ga}
\begin{aligned}
& \mathcal{S}^{(t)} = \text{MUTATION}\big(\mathcal{P}^{(t-1)}\big) \cup \text{CROSSOVER}\big(\mathcal{P}^{(t-1)}\big). \\
& \mathcal{P}^{(t)} = \text{top-K}\big(\mathcal{S}^{(t)}\big). \\
\end{aligned}
} \end{equation}
Specifically, GuacaMol benchmark~\cite{jensen2019graph,brown2019guacamol} proves the superiority of (i) SMILES-based GA and (ii) molecular graph-based GA. 
GA+D~\cite{nigam2019augmenting} leverages SELFIES string as the features of molecule and designs a genetic algorithm enhanced by adversarial learning. 
The superiority of GA methods is also validated in a molecule generation benchmark that optimizes docking score~\cite{huang2021therapeutics}. 

\subsubsection{Monte Carlo Tree Search (MCTS)}
Besides, Monte Carlo Tree Search (MCTS)~\cite{jensen2019graph,yang2020practical} based approaches show its superiority in molecule optimization. 
Different from GA, MCTS can start from scratch ({\eg}, a single node). Then during each generation iteration, it locally and randomly searches each branch of intermediates and selects the most promising ({\eg}, with highest property scores) ones for the next iteration. 
\begin{equation} \small{
\label{eqn:mcts}
\begin{aligned}
& \mathcal{S}^{(t)} = \text{SEARCH-BRANCH}\big(\mathcal{P}^{(t-1)}\big) , \\
& \mathcal{P}^{(t)} = \text{top-K}\big(\mathcal{S}^{(t)}\big), \\
\end{aligned}
} \end{equation}
the process is similar to GA (Equation~\ref{eqn:ga}), where $\mathcal{P}^{(t)}$ is the population at the $t$-th iteration, and $\mathcal{S}^{(t)}$ is the set of all the intermediates generated at the $t$-th iteration.

\subsubsection{Bayesian Optimization (BO)}
Bayesian optimization (BO) is usually used to optimize objective functions that demand a long evaluation time (minutes or hours). 
It is best suited for optimization over continuous domains of less than 20 dimensions~\cite{winter2019efficient} and tolerates stochastic noise in function evaluations. 
BO has two key steps: 
(1) BO builds a surrogate for the objective and quantifies the uncertainty in that surrogate using a Bayesian machine learning technique, \textit{Gaussian process (GP) regression}. Suppose in the $t$-th iteration, we have collected $t$ noisy evaluations, $D_t=\{\mathbf{x}_i, y_i\}_{i=1}^{t}$, where $y_i = f(\mathbf{x}_i) + \epsilon_i$ for iid Gaussian noise $\epsilon_i \in \mathcal{N}(0, \sigma^2)$. 
A GP surrogate model utilizes a non-parametric regression of a particular smoothness controlled by a kernel $k: \mathcal{X} \times \mathcal{X} \xrightarrow[]{} \mathbb{R}$ measuring the similarity between two points.
(2) It utilizes an \textit{acquisition function} defined from the surrogate to decide where to sample.
Specifically, acquisition function $\alpha_t: \mathcal{X} \xrightarrow[]{} \mathbb{R}$ measures the utility of making a new evaluation given the predictions of current surrogate model. Within the $t$-th iteration, BO maximizes the acquisition function via $\mathbf{x}^{(t+1)} = \underset{x\in \mathcal{X}}{\arg\max}\ \alpha_t(x)$. 
Representative BO works include ~\cite{korovina2020chembo,winter2019efficient,notin2021improving,griffiths2020constrained,eriksson2021high,maus2022local}.  
Also, BO methods can be combined VAE and optimize the latent vectors of VAE~\cite{gomez2018automatic,jin2018junction}.  

\subsubsection{Markov Chain Monte Carlo (MCMC)} 
MCMC is designed for drawing samples from target distribution (denoted $\pi(\cdot)$) efficiently. In the $t$-th iteration, MCMC mainly contains two steps: (1) samples the state from proposal distribution, i.e., $x^{(t)} \sim Q(\cdot | x^{(t-1)})$; 
and (2) accepts/rejects the proposal with probability $\min\{1, \frac{\pi(x^{(t)})Q(x^{(t-1)}|x^{(t)})}{\pi(x^{(t-1)})Q(x^{(t)}|x^{(t-1)})}\}$. 

\cite{fu2021mimosa,xie2021mars} formulate the generation task as an sampling problem. To be more concrete, suppose we have $K$ constraints/objectives (denoted as $O_1, \cdots, O_K$) during the optimization procedures, then the target sampling distribution $\pi(\cdot)$ (i.e., the data space of interest) is formulated as 
\begin{equation} \small{
\pi(Y) \propto \prod_{k=1}^{K} O_k(Y), 
} \end{equation}
where $Y$ is a valid data (like chemically valid molecule). Usually MCMC utilizes a pretrained proposal distribution for sampling (usually a pretrained Graph Neural Network), {\ie}, $Q(Y^{(t+1)} | Y^{(t)})$, where $Y^{(t)}$ is the data at the $t$-th iteration.

\subsection{Other Approaches}

\textbf{Optimal Transport (OT)} is introduced when matching between two groups of molecules, especially for the molecule conformation generation problem where each molecule has multiple associated conformations and OT provides a way to optimize the generative model via measuring the distance between two probability distributions. 

\noindent\textbf{Differentiable Learning} formulate discrete molecules into differentiable formats and enable gradient-based continuous optimization on the molecules directly. DST~\cite{fu2021differentiable} formulates it by directly optimizing through the graph representation of molecules.  



%% file: 04_tasks.tex
\begin{table*}[h]
\centering
\caption{
The input, output, and the goal that we are interested in w.r.t. the generation phase. NA means the task is not meaningful. 
}
\label{tab:task2}
\begin{adjustbox}{max width=\textwidth}
\begin{tabular}{l l l l l}
\toprule
Input, Output & goal & 1D String & \makecell[l]{2D Graph\\(small molecule)} & \makecell[l]{3D Geometry\\(small molecule)} \\
\midrule

\multirow{2}{*}{None}
& goal-free & \textit{de novo} 1D molecule generation & \firsttask & \seventhtask \\
& goal-oriented & \textit{de novo} 1D molecule optimization & \sixthtask & \textit{de novo} \eighthtask \\
\midrule

\multirow{2}{*}{1D String}
& goal-free & NA & NA & \secondtask\\
& goal-oriented & 1D molecule optimization & NA & NA\\
\midrule

\multirow{2}{*}{2D Graph (small molecule)}
& goal-free & NA & NA & \secondtask \\
& goal-oriented & NA & \fifthtask & \textit{de novo} \eighthtask \\
\midrule

\multirow{2}{*}{3D Geometry (small molecule)}
& goal-free & NA & NA & NA\\
& goal-oriented & NA & NA & \eighthtask \\
\midrule

\multirow{2}{*}{3D Geometry (large molecule)}
& goal-free & NA & NA & NA\\
& goal-oriented & NA & NA & \thirdtask\\
\midrule

\multirow{2}{*}{\makecell[l]{1D String/2D Graph (small molecule) +\\3D Geometry (large molecule)}}
& goal-free & NA & NA & NA\\
& goal-oriented & NA & NA & \fourthtask \\
\bottomrule
\end{tabular}
\end{adjustbox}
\end{table*}

\begin{table*}[]
\caption{2D molecule generation. }
\label{tab:2dmethod}
\centering
\begin{tabular}{l l l l l l l}
\toprule
Generative Task & Model & Input & Output & Generative Method \\ 
\midrule
\multirow{19}{*}{\textit{de novo} 1D/2D molecule generation} 
& SD-VAE~\cite{dai2018syntax} & None & SMILES & VAE \\
& GrammarVAE~\cite{kusner2017grammar} & None & SMILES & VAE \\ 
& ChemVAE~\cite{gomez2018automatic} & None & SMILES & VAE \\ 
& JT-VAE~\cite{jin2018junction} & None & 2D-Graph & VAE \\
& GraphVAE~\cite{simonovsky2018graphvae} & None & 2D-Graph & VAE \\
& CGVAE~\cite{liu2018constrained} & None & 2D-Graph & VAE \\
& SG-VAE~\cite{doi:10.1021/acs.jcim.1c01573} & None & SMILES & VAE\\
\cmidrule(lr){2-5}
& GraphNVP~\cite{madhawa2019graphnvp} & None & 2D-Graph & Flow \\
& MoFlow~\cite{zang2020moflow} & None & 2D-Graph & Flow \\
& GraphAF~\cite{shi2019graphaf} & None & 2D-Graph & Flow+AR \\
& GraphDF~\cite{luo2021graphdf} & None & 2D-Graph & Flow+AR \\
\cmidrule(lr){2-5}
& dcGAN~\cite{bian2019deep} & None & Fingerprint & GAN \\
& Defactor~\cite{assouel2018defactor} & None & 2D-Graph & GAN \\
& ORGAN~\cite{guimaraes2017objective} & None & SMILES & GAN+RL \\
& MolGAN~\cite{cao2018molgan} & None & 2D-Graph & GAN+RL \\
\cmidrule(lr){2-5}
& MolecularRNN~\cite{popova2019molecularrnn} & None & 2D-Graph & AR  \\
& SF-RNN~\cite{flam2021keeping} & None & SELFIES & AR\\
\cmidrule(lr){2-5}
& GraphEBM~\cite{liu2021graphebm} & None & 2D-Graph & EBM \\ 
\midrule
\multirow{21}{*}{\textit{de novo} 1D/2D molecule optimization} 
& RNN-finetune~\cite{segler2018generating} & None & SMILES & AR  \\
\cmidrule(lr){2-5}
& PCVAE~\cite{guo2020property,du2021deep} & None & 2D-Graph & VAE \\
& MDVAE~\cite{du2022interpretable} & None & 2D-Graph & VAE \\
\cmidrule(lr){2-5}
& GB-GA~\cite{jensen2019graph} & None & 2D-Graph & GA \\
& GA+D~\cite{nigam2019augmenting} & None & SELFIES & GA \\ 
& EDM~\cite{kwon2021evolutionary} & None & SMILES & GA \\
& JANUS~\cite{nigam2021janus} & None & SELFIES & GA \\
\cmidrule(lr){2-5}
& MolDQN~\cite{zhou2019optimization} & None & 2D-Graph & RL \\
& RationaleRL~\cite{jin2020multi} & None & 2D-Graph & RL \\
& GCPN~\cite{You2018-xh} & None & 2D-Graph & RL+GAN \\
& REINVENT~\cite{blaschke2020reinvent} & None & SMILES & RL+AR \\ 
& MCMG~\cite{wang2021multi} & None & SMILES & RL+AR \\ 
& GEGL~\cite{ahn2020guiding} & None & 2D-Graph & RL+GA \\
\cmidrule(lr){2-5}
& ChemTS~\cite{yang2017chemts} & None & SMILES & MCTS+AR \\
& MP-MCTS~\cite{yang2020practical} & None & SMILES & MCTS+AR \\
\cmidrule(lr){2-5}
& MIMOSA~\cite{fu2021mimosa} & None & 2D-Graph & MCMC \\ 
& MARS~\cite{xie2021mars} & None & 2D-Graph & MCMC \\
\hline
\multirow{20}{*}{1D/2D molecule generation}
& TransVAE~\cite{dollar2021attention} & SMILES & SMILES & VAE \\
& ScaffoldMD~\cite{lim2020scaffold} & 2D-Graph & 2D-Graph & VAE \\
& Hier-VAE~\cite{pmlr-v119-jin20a} & 2D-Graph & 2D-Graph & VAE \\
& NeVAE~\cite{samanta2020nevae} & 2D-Graph+3D-Geometry & 2D-Graph+3D-Geometry & VAE\\
& Cogmol~\cite{chenthamarakshan2020cogmol,das2021accelerated} & Data-agnostic & Data-agnostic & VAE \\ 
& MSO~\cite{winter2019efficient} & SMILES & SMILES & VAE+BO\\
& Constrained-BO-VAE~\cite{griffiths2020constrained} & SMILES & SMILES & VAE+BO \\
& TI-MI/IS-MI~\cite{notin2021improving} & Data-agnostic & Data-agnostic & VAE+BO \\ 
& VJTNN~\cite{jin2018learning} & 2D-Graph & 2D-Graph & VAE+GAN \\ 
& CORE~\cite{fu2020core} & 2D-Graph & 2D-Graph & VAE+GAN \\ 
\cmidrule(lr){2-5}
& Mol-CycleGAN~\cite{maziarka2020mol} & 2D-Graph & 2D-Graph & GAN \\
& LatentGAN~\cite{prykhodko2019novo} & SMILES & SMILES & GAN+AE \\
\cmidrule(lr){2-5}
& FREED~\cite{yang2021hit} & 2D-Graph & 2D-Graph & RL \\
& MOLER~\cite{fu2021moler} & 2D-graph & 2D-graph & RL+VAE \\ 
\cmidrule(lr){2-5}
& STONED~\cite{nigam2021beyond} & SELFIES & SELFIES & GA \\
& SynNet~\cite{gao2021amortized} & 2D-Graph & 2D-Graph & GA \\ 
\cmidrule(lr){2-5}
& QMO~\cite{hoffman2020optimizing} & Data-agnostic & Data-agnostic & Model-agnostic \\
& ChemSpace~\cite{du2022interpreting} & Data-agnostic & Data-agnostic & Model-agnostic \\
\cmidrule(lr){2-5}
& DST~\cite{fu2021differentiable} & 2D-Graph & 2D-Graph & Differentiable \\
\bottomrule
\end{tabular}
\end{table*}

\begin{table*}[]
\caption{3D molecule generation.}
\label{tab:3dmethod}
\centering
\begin{tabular}{l l l l l l l}
\toprule
Generative Task & Model & Input & Output & Generative Method \\ 
\midrule
\multirow{12}{*}{\secondtask} & CVGAE~\cite{mansimov2019molecular} & 2D-Graph & 3D-Geometry & VAE \\ 
& GraphDG~\cite{simm2019generative} & 2D-Graph & 3D-Geometry & VAE \\ 
& ConfVAE~\cite{xu2021end} & 2D-Graph & 3D-Geometry & VAE \\ 
& DMCG~\cite{zhu2022direct} & 2D-graph & 3D-Geometry & VAE\\
\cmidrule(lr){2-5}
& ConfGF~\cite{shi2021learning} & 2D-Graph & 3D-Geometry & Diffusion \\ 
& EVFN~\cite{du2021equivariant} & 2D-Graph & 3D-Geometry & Diffusion \\ 
& DGSM~\cite{luo2021predicting} & 2D-Graph & 3D-Geometry & Diffusion \\ 
& GeoDiff~\cite{xu2021geodiff} & 2D-Graph & 3D-Geometry & Diffusion \\ 
\cmidrule(lr){2-5}
& GeoMol~\cite{ganea2021geomol} & 2D-Graph & 3D-Geometry & OT \\ 
\cmidrule(lr){2-5}
& BOKEI~\cite{chan2020bokei} & 2D-Graph & 3D-Geometry & BO \\ 
\midrule 
\multirow{9}{*}{\seventhtask} 
& G-SchNet~\cite{gebauer2019symmetry} & None & 3D-Geometry & AR\\
& GEN3D~\cite{roney2021generating} & None & 2D-Graph+3D-Geometry & AR\\
\cmidrule(lr){2-5}
& MolGym~\cite{simm2020reinforcement} & None & 3D-Geometry & RL \\ 
& COVARIANT~\cite{simm2020symmetry} & None & 3D-Geometry & RL \\
\cmidrule(lr){2-5}
& 3DMolNet~\cite{nesterov3dmolnet} & None & 2D-Graph+3D-Geometry & VAE\\
\cmidrule(lr){2-5}
& E-NFs~\cite{satorras2021n} & None& 3D-Geometry & Flow \\ 
& G-SphereNet~\cite{luo2021autoregressive} & None & 3D-Geometry & Flow+AR\\
\midrule
\multirow{6}{*}{\thirdtask} 
& AutoGrow~\cite{durrant2013autogrow} & None & 3D-Geometry & GA \\ 
\cmidrule(lr){2-5}
& liGAN~\cite{masuda2020generating} & None & 3D-Density & VAE \\
\cmidrule(lr){2-5}
& DeepLigBuilder~\cite{li2021structure} & None & 2D-Graph+3D-Geometry & MCTS+RL \\
\cmidrule(lr){2-5}
& 3DSBDD~\cite{luo20213d} & None & 3D-Geometry & AR \\ 
\midrule
\multirow{3}{*}{\fourthtask} 
& DeepDock~\cite{mendez2021geometric} & 2D-Graph & 3D-Geometry & EBM\\
\cmidrule(lr){2-5}
& EquiBind~\cite{stark2022equibind} & 2D-Graph & 3D-Geometry & OT \\
\midrule
\multirow{5}{*}{\eighthtask}
& BOA~\cite{chan2019bayesian} & 3D-Geometry & 3D-Geometry & BO\\
\cmidrule(lr){2-5}
& 3D-Scaffold~\cite{joshi20213d} & 3D-Geometry & 3D-Geometry & AR\\
& cG-SchNet~\cite{gebauer2022inverse} & 3D-Geometry & 3D-Geometry & AR\\
\cmidrule(lr){2-5}
& Coarse-GrainingVAE~\cite{wang2022generative} & 3D-Geometry & 3D-Geometry & VAE \\

\bottomrule
\end{tabular}
\end{table*}

\section{Molecule Generation Tasks} 
\label{sec:task} 

\subsection{Road Map of Molecule Generation Tasks}
In this section, we identify $8$ distinct molecule generation tasks based on the goal of generation and input/output data representation types: (1) \textit{de novo} 1D/2D molecule generation, (2) \textit{de novo} 1D/2D molecule optimization, (3) 1D/2D molecule optimization, (4) \seventhtask, (5) \secondtask, (6) \thirdtask, (7) \fourthtask, (8) \eighthtask. These tasks are compared in Table~\ref{tab:task2} from various aspects.  
Nevertheless, the eight tasks can be grouped into two categories, 1D/2D molecule generation and 3D molecule generation based on their output data type. To be more concise, we will divide into these two categories, as each sub-task under the same category shares great similarity. It is worth mentioning three concepts before we dive into the details:
\begin{itemize}
\item \textbf{\textit{de novo} v.s. non-\textit{de novo}}, \textit{de novo} denotes that generating a molecule while no reference molecule/lead is given, i.e., generating molecules from scratch. 
\item \textbf{generation v.s. optimization}, generation (a.k.a. \textit{distribution learning}) refers to generating molecules directly or randomly, while optimization (a.k.a. \textit{goal-directed/goal-oriented generation}) refers to generating molecules with optimal or desired properties. 
\item \textbf{molecule v.s. conformation}, molecule refers to the molecule itself represented either in 1D String, 2D Graph or 3D geometry, while conformation specifically refers to the 3D geometry of a molecule.
\end{itemize}

All the generative methods are organized and categorized by different tasks and generative methods in Table~\ref{tab:2dmethod},~\ref{tab:3dmethod}. For most of the existing methods jointly conducted \textit{generation task} and \textit{optimization task}. 
To better distinguish among distinct methods, we put methods under \textit{generation} if their main contributions are distribution learning for molecule design, while we put methods under \textit{optimization} if their main contributions are techniques that handle the molecule optimization task.

\subsection{1D/2D molecule generation}
The goal of 1D/2D molecule generation is to generate novel, diverse, and chemically valid molecules with desired properties. The overall task could be divided into two steps, (1) generating valid molecules, (2) generating/searching/optimizing molecules with desired properties. In the first step, researchers attempt different methods to achieve the goal, from deep generative models to combinatorial-based methods. As mentioned above, we categorize the tasks based on (1) whether it is \textit{de novo} generation, (2) the goal of generation (generation v.s. optimization) in Table~\ref{tab:2dmethod}.


\subsubsection{Problem formulation} 
As mentioned in Section~\ref{sec:representation}, molecules are normally represented by 1D strings and 2D graphs in the 1D/2D molecule generation task. Here we only consider these two representations and formulate molecule generation problems in the following. 


\textbf{1D String representation.} 1D string representation include simplified molecular-input line-entry system (SMILES)~\cite{weininger1988smiles} and self-referencing embedded strings (SELFIES)~\cite{krenn2020self}. SMILES strings need to follow specific chemical rules to guarantee the validity of the molecules. This may lead to the generation of invalid  string~\cite{kusner2017grammar,gomez2018automatic,jin2018junction}. 
SELFIES avoids this issue by augmenting the rules on handling Branch and Ring and can achieve 100\% validness. However, the connection between atoms in a molecule may be transformed into long-term dependencies in string representations, so it is challenging to model them.

\textbf{2D Graph representation.} 2D Graph representation is commonly used for molecule generation due to its ability to capture substructure information. However, as graph objects are usually discrete and high-dimensional (represented by node feature and edge feature matrices), it is nontrivial to design the generative process. There are two common processes, (1) one-shot generation and (2) sequential generation~\cite{zhu2022survey}. One-shot generation refers to generating the node feature and edge feature matrices of a molecular graph at one single step, while sequential generation refers to generating the nodes and edges step by step. 

\subsubsection{\textit{de novo} 1D/2D molecule generation}
The goal of 1D/2D molecule generation is to generate new molecules from the distributions of given datasets or the vast chemical space. 

\noindent\textbf{Representative work}:
Early works on \textit{de novo} 1D/2D molecule generation represent molecules as 1D SMILES strings. 
SMILES-based VAEs (including ChemVAE~\cite{gomez2018automatic}, SD-VAE~\cite{dai2018syntax} GrammarVAE~\cite{kusner2017grammar}) learn a VAE-based model which encodes and decodes SMILES strings and sample new molecules (strings) via decoding from a Gaussian prior. However, the SMILES string does not guarantee the generated molecules to be 100\% valid. This may lead to a poor validity rate of the generated molecules. Later works argue that representing molecules as graphs could provide a better representation, especially for substructures.
GraphNVP~\cite{madhawa2019graphnvp} and GRF~\cite{honda2019graph} first introduce flow-based generative models into molecule generation and design invertible mapping between the input graph and the latent space. However, as both models directly generate node features (i.e. atoms) and edge features (i.e. bonds) without explicit constraints between them, thus the generated molecules are not guaranteed to be valid. GraphAF \cite{shi2019graphaf} instead designs an autoregressive flow-based model to generate a 2D molecular graph sequentially, which adds an atom or a bond each step and guarantees validity via incorporating a valency check in each step.
MoFlow \cite{zang2020moflow} designs a flow-based model to generate molecular graphs in a one-shot manner, which builds the node feature and edge feature matrices as a whole. To ensure validity, it introduces a valency correction mechanism as a post-processing step. The advantage of the one-shot model compared with the autoregressive model is its simplicity and efficiency, and its ability to capture global information. 
However, it scales relatively poorly and requires valency correction as post-processing to guarantee the chemical validity of the generated molecules. 
Nevertheless, both types of flow-based models require dequantization techniques to model discrete molecular graphs with continuous latent variables, which impedes the accuracy. To address it, GraphDF~\cite{luo2021graphdf} proposes a discrete flow-based model, where latent variables are discrete, which greatly saves computation costs and eliminates the negative effect of dequantization.

\subsubsection{\textit{de novo} 1D/2D molecule optimization}
Different from generation tasks that search over the chemical space, the goal of \textit{de novo} 1D/2D molecule optimization is to generate molecules with optimal or desired properties, in the absence of a given molecule. During the optimization process, oracle calls are usually required to provide heuristics for the goal. 

\noindent\textbf{Representative work}: 
Early works directly bias the distribution of the training dataset to generate molecules with desired properties. RNN-finetune~\cite{segler2018generating} first trains the generative model on a large biological activity dataset and fine-tunes it with another specialized dataset. This is applicable in many real-world scenarios when only a small number of desired molecules are available for model training.
GA+D \cite{nigam2019augmenting} designs a SELFIES level genetic algorithm enhanced by a discriminator neural network to search the molecules with optimal properties over the chemical space. 
GB-GA~\cite{jensen2019graph} leverages graph-level genetic algorithm and design mutation and crossover operations with empirical experience to edit the molecule population in each generation (i.e., iteration).
GCPN \cite{You2018-xh} designs a graph convolutional policy network to optimize molecular properties and ensure validity based on a reward function, and generates molecular graphs by adding either an atom or a bond in a single step. 
MolDQN \cite{zhou2019optimization} designs a deep Q-network (a technique for RL method) based on value function learning and generates molecular graphs by either adding an atom/a bond or removing a bond in a single step.
RationaleRL \cite{jin2020multi} collects property-aware substructures as building blocks and leverages reinforcement learning (policy learning) to grow a molecule with desired properties. 
GEGL \cite{ahn2020guiding} mixes genetic algorithm and reinforcement learning by taking the operations of genetic algorithm as basic actions and using reinforcement learning to learn the policy that optimizes the molecular properties. 
MIMOSA and MARS \cite{fu2021mimosa,xie2021mars} formulate molecule optimization as a Bayesian sampling problem, and design graph neural network-based proposal as Markov Chain Monte Carlo (MCMC) kernel. 

\subsubsection{1D/2D molecule optimization}
Different from \textit{de novo} tasks before, the goal of 1D/2D molecule optimization is to generate molecules with optimal or desired properties by optimizing from a given molecule. In many cases, a local search (without changing the input molecule so much) is sufficient and we do not require traversing over the whole chemical space. 

\noindent\textbf{Representative work}: 
VJTNN \cite{jin2018learning} proposes a graph-to-graph translation framework to translate the input molecule (to be optimized) into the new molecule with more desirable properties and similar to the input one. 
It uses a junction tree of molecule substructures as the representation and utilizes a discriminator network to guide the optimization process. 
Within the graph-to-graph translation framework, CORE \cite{fu2020core} designs a copy \& refine mechanism to copy substructures from the input molecule and optimize the generation procedure; MOLER \cite{fu2021moler} leverages a hybrid deep generative model (i.e., graph-to-graph framework) and reinforcement learning (policy gradient); HierVAE \cite{pmlr-v119-jin20a} designs a hierarchical VAE (both encoder and decoder) to model the mapping and capture hierarchical information of molecules during the optimization process. 
MSO \cite{winter2019efficient} represents molecules as SMILES strings and designs a sequence-level VAE-based model to reconstruct the strings from the learned latent space. Then a Bayesian optimization (BO) approach is leveraged to search over the latent space for molecules with desired properties. 
Mol-CycleGAN \cite{maziarka2020mol} designs a cycle-GAN based model that translates input molecules to target molecules with a cycle-consistency loss~\cite{zhu2017unpaired}. 
LatentGAN \cite{prykhodko2019novo} designs a mixture model of autoencoder (AE) and GAN, where AE is used to learn a continuous latent representation of molecules and GAN encourages the generated molecules to be close to the desirable molecules.
DST \cite{fu2021differentiable} designs a differentiable molecular graph that is differentiable with respect to the substructure in a molecular graph, thus enabling the gradient-based optimization on the discrete molecular graph for desired properties. ChemSpace~\cite{du2022interpreting} introduces a new perspective that aims to discover smooth latent directions controlling molecular properties. It argues the importance of such latent directions for interpreting the smooth structure changes of molecules corresponding to property changes.

\subsubsection{Data}
Over the past few decades, multiple large-scale databases are curated to advance the field of molecule design~\cite{huang2021therapeutics, du2021graphgt}. Here, we list commonly used datasets in molecule design and the detailed statistics of datasets are available in Table~\ref{tab:dataset}. 

\noindent\textbf{QM9}~\cite{ramakrishnan2014quantum,ruddigkeit2012enumeration} provides quantum chemical properties for a relevant, consistent, and comprehensive chemical space of small organic molecules. 
It contains $134k$ stable small organic molecules with 3D conformations that contain up to $9$ heavy atoms. 

\noindent\textbf{ZINC}~\cite{sterling2015zinc} is a free database of commercially-available compounds for virtual screening. It contains over 230 million purchasable compounds in ready-to-dock, 3D formats. 

\noindent\textbf{Molecular Sets (MOSES)}~\cite{polykovskiy2020molecular} is a benchmark platform that provides a cleaned dataset of molecules that are ideal for optimization. It is processed from the ZINC Clean Leads dataset and contains around 1.9 M molecules. 

\noindent\textbf{ChEMBL}~\cite{gaulton2012chembl} is a manually curated database of bioactive molecules with drug-like properties. It brings together chemical, bioactivity, and genomic data to aid the translation of genomic information into effective new drugs. It contains 2.0M molecules.

\noindent\textbf{CEPDB}~\cite{hachmann2011harvard} contains around 4.3 M molecules that are related to organic solar cell materials for the design of organic Photovoltaics. 

\noindent\textbf{GDB13}~\cite{blum2009970} enumerates small organic molecules containing up to 13 atoms of C, N, O, S, and Cl following simple chemical stability and synthetic feasibility rules. It is the largest publicly available small organic molecule database to date and contains around 970 M molecules. 


\subsection{3D molecule generation}
3D structures of molecules are critical to their functions with a wide range of applications such as molecular dynamics, docking, etc. In 3D molecule generation, the goal is to generate molecules in 3D space. Unlike 1D/2D molecule generation, one 1D/2D molecule has a variety of 3D geometry or conformations. This results in a list of tasks in which the final outputs are expected to be 3D molecules. As mentioned above, we categorize the tasks based on (1) whether it is \textit{de novo} generation, (2) the goal of generation (generation v.s. optimization), and (3) the input/output of the generation (2D molecule v.s. 3D conformation) in Table~\ref{tab:3dmethod}. Similar to 1D/2D molecule generation, two key challenges also exist for 3D molecule generation: (1) generating both syntactically- and geometrically-valid molecules, (2) optimizing molecule structures with desired properties. We will introduce each task under 3D molecule generation in the following sections.


\subsubsection{Problem formulation}
As mentioned in Section~\ref{sec:representation}, the most common 3D representation of molecules is 3D geometry which consists of a set of nodes with 3D coordinates. 3D representations also include density map, surface mesh, etc.. Here we mainly consider 3D geometry representation and briefly explain other representations in the correspondingly related works.

\textbf{Geometry representation.} Geometry representation is a natural representation of molecules characterized by a set of nodes and coordinates in 3D space. One main challenge in representing molecules as geometry is that the model needs to be equivariant to rotation and translation in 3D space. This needs to be properly addressed in reducing the sample complexity of learning the generative methods.

\subsubsection{\seventhtask}
The goal of \text{\seventhtask} is to generate novel, diverse, and chemically \& geometrically valid molecules from scratch. Compared to \firsttask, this task is more flexible to generate molecules in 3D space, while it is also more challenging since it has to take into account geometric validity beyond syntactic validity. While the additional benefit of generating in 3D is that it could be utilized to take into account the geometric constraint or binding conformation where 3D geometry is needed.

\noindent\textbf{Representative work}: Early attempts focus on extending the generative methods in \firsttask, with 3D geometric representation learning. G-SchNet~\cite{gebauer2019symmetry} argues that graph-based generative models have natural flaws in recognizing spatial isomerism and non-bonded interactions of molecules. Specifically, it designs a purely 3D-based autoregressive model which eliminates the need for representing molecules as graphs and respects the rotational invariance of the molecule structures. 3DMolNet~\cite{nesterov3dmolnet} designs a VAE-based generative model instead of the autoregressive model which learns a latent space for smooth exploration of the chemical space. E-NFs~\cite{satorras2021n} also respects the symmetry via integrating an equivariant graph neural network in a normalizing flow-based generative model. To obtain an invertible equivariant function for normalizing flow, it integrates the equivariant graph neural network in neural ordinary differential equations which are continuous-time differentiable.  G-SphereNet~\cite{luo2021autoregressive} extends the idea of G-SchNet to design an autoregressive model which places atoms in 3D space step-by-step with the backbone model to be one recent 3D GNN, SphereNet~\cite{liu2021spherical}. Specifically, the 3D geometry is generated via generating distances, angles, and torsion angles which determine the positions of each atom rather than generating coordinates directly.

\subsubsection{\secondtask}
The goal of \text{\secondtask} is to generate novel, diverse, and chemically \& geometrically valid molecule conformations from given 2D molecules. Generating 3D conformations from 2D molecules is of great importance to many downstream tasks such as pocket-ligand binding affinity prediction, pocket-based molecule design, etc. 

\noindent\textbf{Representative work}: 
Early work CVGAE \cite{mansimov2019molecular} designs a conditional VAE-based model that learns to generate 3D conformations given the 2D molecules as conditions. It directly generates the 3D geometry. However, a post-alignment, which aligns the generated conformations to the ground truth conformations, is needed to ensure the translation- and rotation-equivariance of the model.
Instead of directly generating the 3D geometry, ConfDG \cite{simm2019generative} combines a conditional VAE that generates the interatomic distances over molecule atoms with a euclidean distance geometry which translates the generated distances to a set of atomic coordinates.
Later, ConfVAE~\cite{xu2021end} develops an end-to-end framework that combines conditional VAE and distance geometry and eliminates the need to separate these two steps. 
GeoMol~\cite{ganea2021geomol} presents a generative learning pipeline for this task which first predicts local atomic 3D structures and torsion angles and then assembles the conformations, to avoid unnecessary over-parameterization of the geometric degrees of freedom. 
Another popular generative models used in this task is score-based or diffusion generative models. ConfGF~\cite{shi2021learning} first devises a score-based generative model to estimate the gradient fields of log density with respect to interatomic distances and calculates the gradient fields of log density of atomic coordinates via the chain rule. Then it leverages Langevin dynamics for efficient sampling of molecule conformations. DGSM~\cite{luo2021predicting} extends it to take into account the non-bonded interactions by making the molecular graph structure learnable~\cite{zhusurvey}. 
EVFN~\cite{du2021equivariant} designs an Equivariant Vector Field Network (EVFN) (with a novel tuple of equivariant basis and the associated scalarization and vectorization layers) to learn gradient vector fields that allow direct modeling of the 3D coordinates of the conformations via score-based generative models. 
GeoDiff~\cite{xu2021geodiff} also designs a diffusion model directly over the 3D geometries instead of interatomic distances to estimate the gradient of log-probability of conformations and sampling from the probability. 
DMCG~\cite{zhu2022direct} realizes the importance of permutation-, translation- and rotation-equivaraince in  directly generating 3D geometry from 2D molecular graphs. It introduces an alignment and permutation post-processing step each time before calculating the loss between the generated and ground truth geometries to relieve the model from learning these equivariance properties in the training phase.

\subsubsection{\thirdtask}
The goal of \text{\thirdtask} is to design 3D molecules given proteins' binding pockets. The designed 3D molecules are expected to have high binding affinities to the protein targets. As mentioned above, designing molecules that could bind to specific pockets is a critical task in structured-based drug discovery. 

\noindent\textbf{Representative work}: liGAN~\cite{masuda2020generating} proposes to represent 3D molecules by 3D atomic density maps which are more smooth than the discrete graph structures and devises a conditional VAE-based approach with 3D CNN as backbone models. However, as the density map is not a direct representation of molecules, it takes further processes to fit the atoms into the density map and make bonds based on chemical rules (e.g., bond length, angle, atom type) after generating the atomic density maps. DeepLigBuilder~\cite{li2021structure} instead represents molecules in 3D geometries which are constructed by 3D coordinates and atom types. Specifically, it designs a RL-based 3D molecule generative model which has a policy network to iteratively refine existing structure and add new atoms to the structure. Furthermore, an MCTS algorithm is combined with the RL method to perform structure-based drug discovery towards a specific protein binding pocket. 3DSBDD~\cite{luo20213d} also represents molecules by 3D geometry and designs an autoregressive model to sample new molecules with high probability in the geometric space of the specific protein binding pocket.

\subsubsection{\fourthtask}
Different from \text{\secondtask} which aims to generate a diverse set of molecule conformations without specific targets, the goal of \text{\fourthtask} is to find the appropriate conformation of an input (given) molecule to the given protein's pocket site. 

\noindent\textbf{Representative work}: DeepDock~\cite{mendez2021geometric} designs a model that takes 2D molecular graphs and 2D protein pocket graphs as input and learns the continuous representations. Then, the model learns a statistical potential based on distance likelihood for the pairwise protein-molecule representations. Finally, an optimization method is performed to produce the binding conformation of the molecule. EquiBind~\cite{stark2022equibind} instead represents molecules and protein pockets with 3D geometries and models them with a 3D equivariant graph neural network. Then, the binding pose conformation of the molecule towards the target is optimized through a learned set of key points matching from the attention map of a graph matching neural network. 

\subsubsection{\eighthtask}
The goal of \text{\eighthtask} is to optimize the molecule represented in 3D to meet desired requirements, e.g., better binding pose or better molecular property. Note that although high binding affinity could be considered as one of the goals for this task, we separate related work to \text{\fourthtask} as it is important to be a stand-alone task. Here, we include a diverse set of \text{\eighthtask} tasks, such as low-energy molecule conformation optimization, scaffold-based 3D molecule optimization, coarse-to-fine molecule conformation optimization.

\noindent\textbf{Representative work}: 
BOA~\cite{chan2019bayesian} searches for molecule conformations with the lowest energy which are most probable and stable via Bayesian Optimization. 
3D-Scaffold~\cite{joshi20213d} focuses on generating 3D molecules with the replacement of the 3D scaffolds. Specifically, it builds upon G-SchNet and autoregressively generates new atoms from fixed existing 3D scaffolds and the generated new molecule candidates are guaranteed to have the desired scaffolds.
Coarse-GrainingVAE~\cite{wang2022generative} learns to generate fine-grained molecule conformations from the coarse-grained molecule conformations via a VAE-based generative model that reconstructs/samples the fine-grained conformations.

\subsubsection{Data}
3D molecule and conformation databases have also been accumulated in the past few decades with the rising of multiple large data repositories such as Protein Data Bank (PDB)~\cite{berman2000protein}. We list the main datasets used for 3D molecule generation in the following. The detailed statistics of datasets are available in Table~\ref{tab:dataset}. 

\noindent\textbf{GEOM-QM9}~\cite{axelrod2020geom} uses advanced sampling and semi-empirical density functional theory to annotate the 3D conformer ensembles. It contains around 133,000 3D molecules. 

\noindent\textbf{GEOM-Drugs}~\cite{axelrod2020geom} also uses advanced sampling and semi-empirical density functional theory to annotate the 3D conformer ensembles. It contains around 317,000 3D molecules. 

\noindent\textbf{ISO17}~\cite{schutt2017quantum} contains 197 2D molecules and 430,692 molecule-conformation pairs. 

\noindent\textbf{Molecule3D}~\cite{xu2021molecule3d} is a dataset with precise ground-state geometries of approximately 4 million molecules derived
from density functional theory (DFT).

\noindent\textbf{CrossDock2020}~\cite{francoeur2020three} contains a total of 22,500,000 poses of ligands docked into multiple similar binding pockets across the Protein Data Bank. 

\noindent\textbf{scPDB}~\cite{desaphy2015sc} is a comprehensive and up-to-date selection of ligandable binding sites of the Protein Data Bank. It has a total of 16,034 entries and registers 9,283 binding sites from 3,678 unique proteins and 5,608 unique ligands. 

\noindent\textbf{DUD-E}~\cite{mysinger2012directory} contains 22,886 active compounds and affinity scores against 102 targets. Each active compound, which has similar physico-chemical properties but different molecules, is associated with 50 decoys.

\begin{table*}[h]
\caption{Available datasets and statistics for various types of molecule design tasks (if both 2D molecules and 3D conformations are available, it is denoted by x$;$y, number in parenthesis refers to large molecule (protein pocket) numbers. }
\label{tab:dataset}
\centering
\begin{tabular}{l l l l l l}
\toprule
Dataset & 1D/2D & 3D & Approx. Amount & Purpose \\
\midrule
ZINC~\cite{irwin2012zinc,sterling2015zinc} & \cmark & & 250,000 & 1D/2D molecule generation\\
ChEMBL~\cite{gaulton2012chembl} & \cmark & & 2,100,000 & 1D/2D molecule generation\\
MOSES~\cite{polykovskiy2020molecular} & \cmark & & 1,937,000 & 1D/2D molecule generation\\
CEPDB~\cite{hachmann2011harvard} & \cmark & & 4,300,000 & 1D/2D molecule generation\\
GDB13~\cite{blum2009970} & \cmark & & 970,000,000 & 1D/2D molecule generation\\
\hline 
QM9~\cite{ramakrishnan2014quantum,ruddigkeit2012enumeration} & \cmark & \cmark & 134,000$;$134,000  & 1D/2D/3D molecule generation \\
GEOM~\cite{axelrod2020geom} & \cmark & \cmark & 450,000$;$37,000,000 & 1D/2D/3D molecule generation \\
ISO17~\cite{schutt2017quantum} & \cmark & \cmark & 200$;$431,000 & 1D/2D/3D molecule generation \\ 
Molecule3D~\cite{xu2021molecule3d} & \cmark & \cmark & 3,900,000$;$3,900,000 & 1D/2D/3D molecule generation \\
CrossDock2020~\cite{francoeur2020three} & \cmark & \cmark & 14,000$;$22,500,000(3,000) & 1D/2D/3D molecule generation \\
scPDB~\cite{desaphy2015sc} & \cmark & \cmark & 6,000$;$16,000(5,000) & 1D/2D/3D molecule generation \\
DUD-E~\cite{mysinger2012directory} & \cmark & \cmark & 23,000$;$1,114,000(100) & 1D/2D/3D molecule generation\\
\bottomrule
\end{tabular}
\end{table*}

%% file: 05_evaluations.tex
\section{Molecule Generation Evaluation} 
\label{sec:evaluation}

The evaluation of molecule generation tasks can be grouped in four categories, (1) evaluations of a molecule generation method, including the validity, novelty, and uniqueness of the generated molecules; (2) distribution evaluations that measure the distribution of generated molecules and distribution of training molecules; (3) optimization evaluations; (4) conformation evaluation.  
Basic evaluations refer to the basic requirement of generating molecules, e.g., validity and geometric stability for 2D and 3D molecule generation. 
Distribution evaluations refer to the distribution of the generated molecules compared to the input distribution. The distribution could be measured by various aspects including molecular properties, graph statistics, sequence similarities, structure similarities, 3D geometry similarities, etc. 
Optimization evaluations highly relate to the goal of the optimization task. For example, molecule optimization task usually aims to generate molecules or optimize molecules towards specific target values. 

However, it is worth noting that not only there are a diverse set of molecules in 3D, but rather one single molecule could have a diverse set of conformations that determines many key properties, especially related to pocket-binding, structured-based drug discovery, etc. In contrast to 1D/2D molecule generation, diversity is not always the desired property for 3D molecule generation. For example, \text{\thirdtask} aims to generate the binding pose or conformation of a molecule for a specific protein pocket. 

\begin{table*}[]
\caption{Evaluation metrics. }
\label{tab:evaluate}
\centering
\begin{tabular}{l l l l l l l}
\toprule
Metric & 1D/2D/3D & Task & Descriptions \\ 
\midrule 
Validity & 1D/2D & generation & percentage of valid molecules among all the generated molecules \\
Novelty & 1D/2D  & generation & percentage of the generated molecules that do not appear in training set \\
Uniqueness & 1D/2D & generation & percentage of unique molecules among all the generated molecules \\
Quality & 1D/2D & generation & percentage of good-quality molecules filtered by chemical rules\\
IntDiversity & 1D/2D & generation & percentage of diverse molecules among all the generated molecules\\
ExtDiversity & 1D/2D & generation & percentage of diverse molecules between training set and generated molecules\\
{Kullback–Leibler Divergence} (KLD) & 1D/2D & generation & distance between training set and generated molecules \\ 
{Fréchet ChemNet Distance} (FCD) & 1D/2D & generation & distance between training set and generated molecules \\
{Mean Maximum Discrepancy} (MMD) & 1D/2D & generation & distance between training set and generated molecules \\
Rediscovery & 1D/2D & generation & rediscovered molecules removed from training set \\
Isomer Discovery & 1D/2D & generation & generated molecules following a simple (unknown) pattern \\
Median Structure Discovery & 1D/2D & generation & generated molecules similar to multiple structures \\
Tanimoto Similarity & 1D/2D & optimization & Tanimoto similarity between training set and generated molecules \\
Scaffold Similarity & 1D/2D & optimization & Bemis-Murcko scaffold similarity between the input and generated molecules\\
Fragment Similarity & 1D/2D & optimization & BRICS fragment similarity between the input and generated molecules\\
{Oracle}  & 1D/2D/3D & optimization & molecular property evaluator \\
Stability & 3D & generation & percentage of atom-level and molecule-level stable molecules\\
{Room-Mean-Square Deviation} (RMSD) & 3D & generation & 3D structure similarity \\
Kabsch-RMSD & 3D & generation & 3D structure similarity after Kabsch superimposition\\
Centroid Distance & 3D & generation & 3D structure distance\\ 
{Coverage} (COV) & 3D & generation & coverage of conformations in training set \\
{Matching} (MAT) & 3D & generation & smallest RMSD between generated conformations and training set \\
\bottomrule
\end{tabular}
\end{table*}

\subsection{Generation Evaluation}
The basic evaluation metrics for molecule generation task include \textbf{Validity}, \textbf{Novelty}, \textbf{Uniqueness}, \textbf{Diversity}~\cite{You2018-xh}. 
\begin{align}
    \text{Validity} = \frac{\text{\# valid molecules}}{\text{\# generated molecules}}\, \\
    \text{Novelty} = \frac{\text{\# novel molecules}}{\text{\# generated molecules}}\, \\
    \text{Uniqueness} = \frac{\text{\# unique molecules}}{\text{\# generated molecules}}\, \\
    \text{Quality} = \frac{\text{\# good molecules}}{\text{\# generated molecules}}\,
\end{align}
where novel molecules are the generated molecules that do not appear in training set. Validity is the percentage of valid molecules among all the generated molecules. 
Novelty is the percentage of the generated molecules that do not appear in training set and measures the ability to explore unknown chemical space. 
Uniqueness is the percentage of unique molecules among all the generated molecules and measures the methods' ability to reconstruct the molecules. 
Quality is the percentage of good molecules filtered by a predefined rule set and measures the ability of the method to generate good molecules.

\textbf{Discovery-based evaluation}~\cite{polykovskiy2020molecular}. Discovery-based evaluation refers to testing whether the generative methods would discover certain patterns or rediscover molecules removed from the training set.


\noindent\textbf{Diversity}~\cite{benhenda2017chemgan} is composed of two parts, internal diversity and external diversity. Internal diversity is defined as the average pairwise similarity among all the generated molecules, while external diversity is defined as the average pairwise similarity between the training set and the generated molecules, as follows:: 
\begin{equation}
\begin{aligned}
\text{Int} = 1 - \frac{1}{|\mathcal{Z}_{gen}|(|\mathcal{Z}_{gen}|-1)}\sum_{Z_1,Z_2 \in \mathcal{Z}_{gen}, Z_1 \neq Z_2} \text{sim}(Z_1,Z_2),\\
\text{Ext} = 1 - \frac{1}{|\mathcal{Z}_{gen}||\mathcal{Z}_{train}|}\sum_{Z_1,Z_2 \in \mathcal{Z}_{gen},\mathcal{Z}_{train}} \text{sim}(Z_1,Z_2),\\
\end{aligned}
\end{equation}
where $\mathcal{Z}$ is the set of generated molecules. $\text{sim}(Z_1,Z_2)$ could be any similarity metric between molecule $Z_1$ and $Z_2$, calculated in different ways, such as Tanimoto similarity, scaffold similarity, fragment similarity, etc. 

\subsection{Distribution Evaluation}

Additionally, more evaluation metrics are used to measure how well the generative models learn the true data property distributions, including 
\begin{itemize}
\item \textbf{Kullback–Leibler Divergence} (KLD)~\cite{kullback1951information}. KL divergence between the probability distributions of a variety of physicochemical descriptors for the training set and a set of generated molecules. Models able to capture the distributions of molecules in the training set will lead to small KL divergence values. 
\item \textbf{Fréchet ChemNet Distance} (FCD)~\cite{preuer2018frechet}. FCD first takes the means and covariances of the activations of the penultimate layer of ChemNet are calculated for the reference set and the set of generated molecules. 
The FCD is then calculated as the Frechet distance for both pairs of values. 
Lower FCD values indicate more similarity between two distributions. 
\item \textbf{Mean Maximum Discrepancy} (MMD)~\cite{gretton2012kernel}. MMD measures the distance between the probability distributions of a variety of physicochemical descriptors for the training set and a set of generated molecules. 
\end{itemize}

\subsection{Optimization Evaluation}

The goal of optimization is to generate molecules with desirable properties. These properties are required to be evaluated. The property evaluator is also named oracle.

\noindent\textbf{Oracle} (a.k.a. molecular property evaluator) can be seen as a black-box function that evaluate certain chemical or biological properties of a molecule. 
It takes a molecule as the input and returns its property. 
The common oracles include 
\begin{itemize}[leftmargin=*]
 \item Synthetic Accessibility (SA) measures how hard to synthesize the given molecule. It is evaluated by RDKit~\cite{landrum2013rdkit}. 
  \item Quantitative Estimate of Drug-likeness (QED) is a quantitative estimate of drug-likeness. It is evaluated by RDKit~\cite{landrum2013rdkit}. 
 \item Octanol-water Partition Coefficient (LogP) measures the solubility and synthetic accessibility of a compound. It is evaluated via RDKit~\cite{landrum2013rdkit}.
  \item Glycogen Synthase Kinase 3 Beta (GSK3$\beta$). The evaluator is a trained random forest classifier using ECFP6 fingerprints using ExCAPE-DB dataset~\cite{li2018multi}. 
 \item c-Jun N-terminal Kinases-3 (JNK3). The evaluator is also a trained random forest classifier using ECFP6 fingerprints using ExCAPE-DB dataset~\cite{li2018multi}. 
 \item Dopamine Receptor D2 (DRD2). The evaluator is a support vector machine (SVM) classifier with a Gaussian kernel with ECFP6 fingerprint on ExCAPE-DB dataset~\cite{olivecrona2017molecular}. 
 \item Guacamol related Oracle. GuacaMol is a popular molecule generation benchmark, which curates lots of oracles, including rediscovery oracle that rediscover molecules removed from training data, Isomer identification oracle that generated molecules that follow a simple (unknown) pattern, and Median molecule discovery that generates molecules that maximize similarity to multiple molecules~\cite{brown2019guacamol}. 
 \item Vina Score. Vina is a scoring function that measures the protein-ligand binding affinity~\cite{trott2010autodock}.
\end{itemize}
QED, GSK3$\beta$, JNK3, and DRD2 scores range from 0 to 1, higher value is more preferred.  
In contrast, the SA score ranges from 0 to 10, lower SA score means the molecule is easier to synthesize and is more desirable. 
JNK3 and GSK3$\beta$ were originally curated in~\cite{li2018multi}. 
DRD2 were curated in~\cite{olivecrona2017molecular}. 

\noindent\textbf{Oracle complexity}. 
In realistic discovery settings, the oracle acquisition cost is usually not negligible. 
Thus, it is necessary to consider oracle complexity as a metric of a molecule generation algorithm. 
\cite{fu2021differentiable} compares the performance of various methods given the same number of oracle calls, because the number of oracle calls is an important metric to evaluate the sample efficiency/complexity of a method. 

\noindent\textbf{Constrained optimization}. In many real-world scenarios, molecule optimization aims to search for molecules with similar structures as the base molecule but with a high activity such as drug-likeness or synthesis accessibility for drug design. Therefore, the similarity between molecule structures is also considered during the so-called constrained optimization scenario. Common similarity measurements are Tanimoto similarity, scaffold similarity, fragment similarity, etc.

\subsection{3D Molecule Evaluation}
This section discusses the evaluation of molecular conformations, measuring the 3D geometries. 

\textbf{Stability}
As first introduced in E-NFs~\cite{satorras2021n}, stability extends from validity in evaluating 1D/2D molecule to its stability in 3D space. Specifically, an atom is stable when the number of bonds connecting to other atoms matches the valence of the atom, a molecule is stable when all atoms are stable. 

\textbf{Room-Mean-Square Deviation} (RMSD) measures the alignment between two conformations $R\in \mathbb{R}^{n\times 3}$ and $\hat{R}\in \mathbb{R}^{n\times 3}$, $n$ is the number of points in the conformation, defined as follows:
\begin{equation}
 \text{RMSD}(R,\hat{R}) = \big(\frac{1}{n}\sum_{i=1}^{n}||R_i - \hat{R}_i||^2 \big)^{\frac{1}{2}}.
\end{equation}
The conformation $\hat{R}$ is obtained by an alignment function $\hat{R} = A(R, R^{r})$, which rotates and translates the reference conformation $R^r$ to have the smallest distance to the generated R according to the RMSD metrics. Kabsch-RMSD is a type of RMSDs that first superimposes the two structures with the Kabsch algorithm before calculating RMSD scores.

Two common metrics to evaluate the quality of the generated molecule conformations are \textbf{Coverage} (COV) and \textbf{Matching} (MAT) scores~\cite{xu2021geodiff,luo20213d}. 
\begin{equation}
\text{COV}(S_g, S_r) = \frac{1}{|S_r|}|\{R \in S_r|\text{RMSD}(R,\hat{R})<\delta, \hat{R} \in S_g\}|, 
\end{equation}
\begin{equation}
\text{MAT}(S_g, S_r) = \frac{1}{|S_r|}\sum_{R \in S_r} \min_{\hat{R} \in S_g} \text{RMSD}(R, \hat{R}),
\end{equation}
where $S_g$ and $S_r$ denote generated and reference conformations, respectively. $\delta$ is a given \textbf{RMSD} threshold.

%% file: 06_future_directions.tex
\section{Future Direction}
\label{sec:future}

This section summarizes challenges we face and the opportunities we have in molecule generation from various aspects, including data, oracle, interpretability, evaluation, prior knowledge, tasks, etc. 

\subsection{Challenges}
\begin{itemize}
\item \textbf{Out-of-distribution Generation}, the known molecules take up a small fraction of all the chemically valid molecules. Most of the existing methods (especially deep generative models) imitate the known molecule distribution and lack the exploration of the unknown molecule space. 
\item \textbf{Unrealistic Problem Formulation}, the problem formulation for molecule design task should respect real-world applications in chemistry. 
\item \textbf{Ideal assumptions/expensive Oracle Calls}, current machine learning methods usually assume unlimited access to oracle calls which are expensive in the real-world drug discovery process.
\item \textbf{Lack of Interpretability}, interpretability is an important yet not a fully explored area for molecule generative models. For example, how the generative or optimization process of molecules could lead to interpretable chemical rules. There are some early explorations to the interpretability, e.g., ~\cite{jin2020multi,guo2021deep,fu2021differentiable,du2022interpretable,du2022interpreting,du2022disentangled}. However, most of these methods analyze interpretability via case studies, how to quantitatively evaluate interpretability remains a challenge. 
\item \textbf{Lack of Unified Evaluation Protocols}, this is also related to how we define a good drug candidate, from this, we may also focus more on e.g. structured-based drug design, to generate ligands that could be bound to protein pockets. 
\item \textbf{Lack of Large-scale Study and Benchmark}, tons of machine learning methods have been developed while there are no fair benchmark results on many types of models in different key tasks. The chemical space is estimated to have from $10^{23}$ to $10^{60}$ drug-like molecules, while the current available large datasets, e.g., GDB13, have not been utilized to facilitate molecule generation. Only small fractions of large databases are utilized instead.
\item \textbf{Generation in Low-data Regime}, despite the current success in generating new molecules with machine learning models learned from large molecule data repositories, many real-world applications are in short of data collections. Thus, generating molecules in low-data regime remains a challenge that limits the real-world applications of molecule design.
\end{itemize}
\subsection{Opportunities}
\begin{itemize}
\item \textbf{Extend to more structured data}. Essentially, this survey aims at giving a comprehensive description of small molecule generation, which is indeed structured data. Yet, other applications with more complex data structures may also benefit from the approaches discussed here, {\eg}, the design of polysaccharide, protein, antibody, gene, crystal structure, etc.
\item \textbf{Connection to later phases in drug development}. Molecule design/optimization are early phases in drug discovery. Current approaches leverage some simple and ad-hoc oracles, so there is a gap between molecule design/optimization and later phases in drug development (such as animal models (a.k.a. pre-clinical) and clinical trials)~\cite{fu2022hint,fu2021probabilistic}. It would be helpful if we can leverage the information from these later phases.  
\item \textbf{Knowledge Discovery}. Modeling chemical compounds in an efficient and structured way could lead to enhanced human knowledge. Graph structure learning could infer the implicit non-bonded or long-range interactions which are usually ignored while modeling molecules with chemical laws. The exploration over the latent space could interpret the chemical rules that govern molecular properties. 
\end{itemize}

%% file: 07_conclusion.tex
\section{Conclusion}
\label{sec:conclusion}

In this survey, we comprehensively review the molecule design problem under the setting of machine learning. To the best of our knowledge, this is the first comprehensive review of machine learning approaches for molecule design. We categorize the generation methods on molecules from mainly two aspects. Furthermore, we elaborate on the molecule design tasks and empirical setups of each individual task. Finally, we identify open challenges and opportunities in molecule design, point out the limitations of current approaches and pave the way for the future of molecule design with machine learning models.

%% file: main.bbl
\begin{thebibliography}{100}
\providecommand{\url}[1]{#1}
\csname url@samestyle\endcsname
\providecommand{\newblock}{\relax}
\providecommand{\bibinfo}[2]{#2}
\providecommand{\BIBentrySTDinterwordspacing}{\spaceskip=0pt\relax}
\providecommand{\BIBentryALTinterwordstretchfactor}{4}
\providecommand{\BIBentryALTinterwordspacing}{\spaceskip=\fontdimen2\font plus
\BIBentryALTinterwordstretchfactor\fontdimen3\font minus
  \fontdimen4\font\relax}
\providecommand{\BIBforeignlanguage}[2]{{%
\expandafter\ifx\csname l@#1\endcsname\relax
\typeout{** WARNING: IEEEtran.bst: No hyphenation pattern has been}%
\typeout{** loaded for the language `#1'. Using the pattern for}%
\typeout{** the default language instead.}%
\else
\language=\csname l@#1\endcsname
\fi
#2}}
\providecommand{\BIBdecl}{\relax}
\BIBdecl

\bibitem{drews2000drug}
J.~Drews, ``Drug discovery: a historical perspective,'' \emph{science}, vol.
  287, no. 5460, pp. 1960--1964, 2000.

\bibitem{dimasi2016innovation}
J.~A. DiMasi, H.~G. Grabowski, and R.~W. Hansen, ``Innovation in the
  pharmaceutical industry: new estimates of r\&d costs,'' \emph{Journal of
  health economics}, vol.~47, pp. 20--33, 2016.

\bibitem{wouters2020estimated}
O.~J. Wouters, M.~McKee, and J.~Luyten, ``Estimated research and development
  investment needed to bring a new medicine to market, 2009-2018,''
  \emph{Jama}, vol. 323, no.~9, pp. 844--853, 2020.

\bibitem{polishchuk2013estimation}
P.~G. Polishchuk, T.~I. Madzhidov, and A.~Varnek, ``Estimation of the size of
  drug-like chemical space based on gdb-17 data,'' \emph{Journal of
  computer-aided molecular design}, vol.~27, no.~8, pp. 675--679, 2013.

\bibitem{gomez2018automatic}
R.~G{\'o}mez-Bombarelli, J.~N. Wei, D.~Duvenaud, J.~M. Hern{\'a}ndez-Lobato,
  B.~S{\'a}nchez-Lengeling, D.~Sheberla, J.~Aguilera-Iparraguirre, T.~D.
  Hirzel, R.~P. Adams, and A.~Aspuru-Guzik, ``Automatic chemical design using a
  data-driven continuous representation of molecules,'' \emph{ACS central
  science}, 2018.

\bibitem{nigam2019augmenting}
A.~Nigam, P.~Friederich, M.~Krenn, and A.~Aspuru-Guzik, ``Augmenting genetic
  algorithms with deep neural networks for exploring the chemical space,'' in
  \emph{ICLR}, 2020.

\bibitem{bronstein2017geometric}
M.~M. Bronstein, J.~Bruna, Y.~LeCun, A.~Szlam, and P.~Vandergheynst,
  ``Geometric deep learning: going beyond euclidean data,'' \emph{IEEE Signal
  Processing Magazine}, vol.~34, no.~4, pp. 18--42, 2017.

\bibitem{sanchez2018inverse}
B.~Sanchez-Lengeling and A.~Aspuru-Guzik, ``Inverse molecular design using
  machine learning: Generative models for matter engineering,'' \emph{Science},
  vol. 361, no. 6400, pp. 360--365, 2018.

\bibitem{xue2019advances}
D.~Xue, Y.~Gong, Z.~Yang, G.~Chuai, S.~Qu, A.~Shen, J.~Yu, and Q.~Liu,
  ``Advances and challenges in deep generative models for de novo molecule
  generation,'' \emph{Wiley Interdisciplinary Reviews: Computational Molecular
  Science}, vol.~9, no.~3, p. e1395, 2019.

\bibitem{elton2019deep}
D.~C. Elton, Z.~Boukouvalas, M.~D. Fuge, and P.~W. Chung, ``Deep learning for
  molecular design—a review of the state of the art,'' \emph{Molecular
  Systems Design \& Engineering}, vol.~4, no.~4, pp. 828--849, 2019.

\bibitem{vanhaelen2020advent}
Q.~Vanhaelen, Y.-C. Lin, and A.~Zhavoronkov, ``The advent of generative
  chemistry,'' \emph{ACS Medicinal Chemistry Letters}, vol.~11, no.~8, pp.
  1496--1505, 2020.

\bibitem{alshehri2020deep}
A.~S. Alshehri, R.~Gani, and F.~You, ``Deep learning and knowledge-based
  methods for computer-aided molecular design—toward a unified approach:
  State-of-the-art and future directions,'' \emph{Computers \& Chemical
  Engineering}, vol. 141, p. 107005, 2020.

\bibitem{jimenez2020drug}
J.~Jim{\'e}nez-Luna, F.~Grisoni, and G.~Schneider, ``Drug discovery with
  explainable artificial intelligence,'' \emph{Nature Machine Intelligence},
  vol.~2, no.~10, pp. 573--584, 2020.

\bibitem{axelrod2022learning}
S.~Axelrod, D.~Schwalbe-Koda, S.~Mohapatra, J.~Damewood, K.~P. Greenman, and
  R.~G{\'o}mez-Bombarelli, ``Learning matter: Materials design with machine
  learning and atomistic simulations,'' \emph{Accounts of Materials Research},
  2022.

\bibitem{weininger1988smiles}
D.~Weininger, ``{SMILES}, a chemical language and information system. 1.
  introduction to methodology and encoding rules,'' \emph{Journal of chemical
  information and computer sciences}, vol.~28, no.~1, pp. 31--36, 1988.

\bibitem{krenn2020self}
M.~Krenn, F.~H{\"a}se, A.~Nigam, P.~Friederich, and A.~Aspuru-Guzik,
  ``Self-referencing embedded strings ({SELFIES}): A 100\% robust molecular
  string representation,'' \emph{Machine Learning: Science and Technology},
  vol.~1, no.~4, p. 045024, 2020.

\bibitem{heller2015inchi}
S.~R. Heller, A.~McNaught, I.~Pletnev, S.~Stein, and D.~Tchekhovskoi,
  ``{InChI}, the {IUPAC} international chemical identifier,'' \emph{Journal of
  cheminformatics}, vol.~7, no.~1, pp. 1--34, 2015.

\bibitem{smarts}
\BIBentryALTinterwordspacing
N.~Santa~Fe, ``Daylight chemical information systems, inc.'' [Online].
  Available: \url{www.daylight.com/dayhtml/doc/theory/theory.smarts.html}
\BIBentrySTDinterwordspacing

\bibitem{hirohara2018convolutional}
M.~Hirohara, Y.~Saito, Y.~Koda, K.~Sato, and Y.~Sakakibara, ``Convolutional
  neural network based on {SMILES} representation of compounds for detecting
  chemical motif,'' \emph{BMC bioinformatics}, vol.~19, no.~19, pp. 83--94,
  2018.

\bibitem{bjerrum2017smiles}
E.~J. Bjerrum, ``{SMILES} enumeration as data augmentation for neural network
  modeling of molecules,'' \emph{arXiv preprint arXiv:1703.07076}, 2017.

\bibitem{liu2018practical}
S.~Liu, M.~Alnammi, S.~S. Ericksen, A.~F. Voter, G.~E. Ananiev, J.~L. Keck,
  F.~M. Hoffmann, S.~A. Wildman, and A.~Gitter, ``Practical model selection for
  prospective virtual screening,'' \emph{Journal of chemical information and
  modeling}, vol.~59, no.~1, pp. 282--293, 2018.

\bibitem{huang2020deeppurpose}
K.~Huang, T.~Fu, L.~M. Glass, M.~Zitnik, C.~Xiao, and J.~Sun, ``{DeepPurpose}:
  A deep learning library for drug-target interaction prediction,''
  \emph{Bioinformatics}, 2020.

\bibitem{honda2019smiles}
S.~Honda, S.~Shi, and H.~R. Ueda, ``{SMILES} transformer: Pre-trained molecular
  fingerprint for low data drug discovery,'' \emph{arXiv preprint
  arXiv:1911.04738}, 2019.

\bibitem{wang2019smiles}
S.~Wang, Y.~Guo, Y.~Wang, H.~Sun, and J.~Huang, ``{SMILES-BERT}: large scale
  unsupervised pre-training for molecular property prediction,'' in
  \emph{Proceedings of the 10th ACM international conference on bioinformatics,
  computational biology and health informatics}, 2019, pp. 429--436.

\bibitem{chithrananda2020chemberta}
S.~Chithrananda, G.~Grand, and B.~Ramsundar, ``{ChemBERTa}: Large-scale
  self-supervised pretraining for molecular property prediction,'' \emph{arXiv
  preprint arXiv:2010.09885}, 2020.

\bibitem{gilmer2017neural}
J.~Gilmer, S.~S. Schoenholz, P.~F. Riley, O.~Vinyals, and G.~E. Dahl, ``Neural
  message passing for quantum chemistry,'' in \emph{International Conference on
  Machine Learning}.\hskip 1em plus 0.5em minus 0.4em\relax PMLR, 2017, pp.
  1263--1272.

\bibitem{kipf2016semi}
T.~N. Kipf and M.~Welling, ``Semi-supervised classification with graph
  convolutional networks,'' \emph{International Conference on Learning
  Representations}, 2016.

\bibitem{hamilton2017inductive}
W.~L. Hamilton, R.~Ying, and J.~Leskovec, ``Inductive representation learning
  on large graphs,'' in \emph{Advances in Neural Information Processing
  Systems, NeurIPS}, 2017.

\bibitem{xu2018powerful}
K.~Xu, W.~Hu, J.~Leskovec, and S.~Jegelka, ``How powerful are graph neural
  networks?'' \emph{International Conference on Learning Representations,
  {ICLR}}, 2018.

\bibitem{duvenaud2015convolutional}
D.~K. Duvenaud, D.~Maclaurin, J.~Iparraguirre, R.~Bombarell, T.~Hirzel,
  A.~Aspuru-Guzik, and R.~P. Adams, ``Convolutional networks on graphs for
  learning molecular fingerprints,'' \emph{Advances in neural information
  processing systems}, vol.~28, 2015.

\bibitem{kearnes2016molecular}
S.~Kearnes, K.~McCloskey, M.~Berndl, V.~Pande, and P.~Riley, ``Molecular graph
  convolutions: moving beyond fingerprints,'' \emph{Journal of computer-aided
  molecular design}, vol.~30, no.~8, pp. 595--608, 2016.

\bibitem{yang2019analyzing}
K.~Yang, K.~Swanson, W.~Jin, C.~Coley, P.~Eiden, H.~Gao, A.~Guzman-Perez,
  T.~Hopper, B.~Kelley, M.~Mathea \emph{et~al.}, ``Analyzing learned molecular
  representations for property prediction,'' \emph{Journal of chemical
  information and modeling}, vol.~59, no.~8, pp. 3370--3388, 2019.

\bibitem{song2020communicative}
Y.~Song, S.~Zheng, Z.~Niu, Z.-H. Fu, Y.~Lu, and Y.~Yang, ``Communicative
  representation learning on attributed molecular graphs.'' in \emph{IJCAI},
  vol. 2020, 2020, pp. 2831--2838.

\bibitem{corso2020principal}
G.~Corso, L.~Cavalleri, D.~Beaini, P.~Li{\`o}, and P.~Veli{\v{c}}kovi{\'c},
  ``Principal neighbourhood aggregation for graph nets,'' \emph{arXiv preprint
  arXiv:2004.05718}, 2020.

\bibitem{liu2018n}
S.~Liu, M.~F. Demirel, and Y.~Liang, ``{N-gram} graph: Simple unsupervised
  representation for graphs, with applications to molecules,'' \emph{Annual
  Conference on Neural Information Processing Systems (NeurIPS)}, pp.
  8464--8476, 2019.

\bibitem{alon2020bottleneck}
U.~Alon and E.~Yahav, ``On the bottleneck of graph neural networks and its
  practical implications,'' \emph{arXiv preprint arXiv:2006.05205}, 2020.

\bibitem{vaswani2017attention}
A.~Vaswani, N.~Shazeer, N.~Parmar, J.~Uszkoreit, L.~Jones, A.~N. Gomez,
  {\L}.~Kaiser, and I.~Polosukhin, ``Attention is all you need,'' in
  \emph{NIPS}, 2017, pp. 5998--6008.

\bibitem{devlin2018bert}
J.~Devlin, M.~Chang, K.~Lee, and K.~Toutanova, ``{BERT:} pre-training of deep
  bidirectional transformers for language understanding,'' in \emph{Proceedings
  of the 2019 Conference of the North American Chapter of the Association for
  Computational Linguistics: Human Language Technologies, {NAACL-HLT}
  2019.}\hskip 1em plus 0.5em minus 0.4em\relax Association for Computational
  Linguistics, 2019, pp. 4171--4186.

\bibitem{dosovitskiy2020image}
A.~Dosovitskiy, L.~Beyer, A.~Kolesnikov, D.~Weissenborn, X.~Zhai,
  T.~Unterthiner, M.~Dehghani, M.~Minderer, G.~Heigold, S.~Gelly \emph{et~al.},
  ``An image is worth 16x16 words: Transformers for image recognition at
  scale,'' \emph{arXiv preprint arXiv:2010.11929}, 2020.

\bibitem{velivckovic2017graph}
P.~Veli{\v{c}}kovi{\'c}, G.~Cucurull, A.~Casanova, A.~Romero, P.~Lio, and
  Y.~Bengio, ``Graph attention networks,'' \emph{International Conference on
  Learning Representations, {ICLR}}, 2017.

\bibitem{xiong2019pushing}
Z.~Xiong, D.~Wang, X.~Liu, F.~Zhong, X.~Wan, X.~Li, Z.~Li, X.~Luo, K.~Chen,
  H.~Jiang \emph{et~al.}, ``Pushing the boundaries of molecular representation
  for drug discovery with the graph attention mechanism,'' \emph{Journal of
  medicinal chemistry}, vol.~63, no.~16, pp. 8749--8760, 2019.

\bibitem{rong2020self}
Y.~Rong, Y.~Bian, T.~Xu, W.~Xie, Y.~Wei, W.~Huang, and J.~Huang,
  ``Self-supervised graph transformer on large-scale molecular data,'' in
  \emph{Advances in Neural Information Processing Systems, NeurIPS}, 2020.

\bibitem{demirel2021analysis}
M.~F. Demirel, S.~Liu, S.~Garg, and Y.~Liang, ``An analysis of attentive
  walk-aggregating graph neural networks,'' \emph{arXiv preprint
  arXiv:2110.02667}, 2021.

\bibitem{ying2021transformers}
C.~Ying, T.~Cai, S.~Luo, S.~Zheng, G.~Ke, D.~He, Y.~Shen, and T.-Y. Liu, ``Do
  transformers really perform badly for graph representation?'' \emph{Advances
  in Neural Information Processing Systems}, vol.~34, 2021.

\bibitem{axelrod2020geom}
S.~Axelrod and R.~Gomez-Bombarelli, ``{GEOM}: Energy-annotated molecular
  conformations for property prediction and molecular generation,'' \emph{arXiv
  preprint arXiv:2006.05531}, 2020.

\bibitem{moss1996basic}
G.~P. Moss, ``Basic terminology of stereochemistry ({IUPAC} recommendations
  1996),'' \emph{Pure and applied chemistry}, vol.~68, no.~12, pp. 2193--2222,
  1996.

\bibitem{schutt2018schnet}
K.~T. Sch{\"u}tt, H.~E. Sauceda, P.-J. Kindermans, A.~Tkatchenko, and K.-R.
  M{\"u}ller, ``Schnet--a deep learning architecture for molecules and
  materials,'' \emph{The Journal of Chemical Physics}, vol. 148, no.~24, p.
  241722, 2018.

\bibitem{qiao2020orbnet}
Z.~Qiao, M.~Welborn, A.~Anandkumar, F.~R. Manby, and T.~F. Miller~III,
  ``Orbnet: Deep learning for quantum chemistry using symmetry-adapted
  atomic-orbital features,'' \emph{The Journal of chemical physics}, vol. 153,
  no.~12, p. 124111, 2020.

\bibitem{klicpera2020directional}
J.~Klicpera, J.~Gro{\ss}, and S.~G{\"u}nnemann, ``Directional message passing
  for molecular graphs,'' \emph{arXiv preprint arXiv:2003.03123}, 2020.

\bibitem{klicpera2020fast}
J.~Klicpera, S.~Giri, J.~T. Margraf, and S.~G{\"u}nnemann, ``Fast and
  uncertainty-aware directional message passing for non-equilibrium
  molecules,'' \emph{arXiv preprint arXiv:2011.14115}, 2020.

\bibitem{fuchs2020se}
F.~B. Fuchs, D.~E. Worrall, V.~Fischer, and M.~Welling, ``Se(3)-transformers:
  3d roto-translation equivariant attention networks,'' \emph{Annual Conference
  on Neural Information Processing Systems (NeurIPS)}, 2020.

\bibitem{satorras2021n}
V.~G. Satorras, E.~Hoogeboom, and M.~Welling, ``E(n) equivariant graph neural
  networks,'' \emph{Proceedings of the 38th International Conference on Machine
  Learning, {ICML}}, vol. 139, pp. 9323--9332, 2021.

\bibitem{shuaibi2021rotation}
M.~Shuaibi, A.~Kolluru, A.~Das, A.~Grover, A.~Sriram, Z.~Ulissi, and C.~L.
  Zitnick, ``Rotation invariant graph neural networks using spin
  convolutions,'' \emph{arXiv preprint arXiv:2106.09575}, 2021.

\bibitem{liu2021spherical}
Y.~Liu, L.~Wang, M.~Liu, X.~Zhang, B.~Oztekin, and S.~Ji, ``Spherical message
  passing for 3d graph networks,'' \emph{arXiv preprint arXiv:2102.05013},
  2021.

\bibitem{du2021equivariant}
W.~Du, H.~Zhang, Y.~Du, Q.~Meng, W.~Chen, B.~Shao, and T.-Y. Liu, ``Equivariant
  vector field network for many-body system modeling,'' \emph{arXiv preprint
  arXiv:2110.14811}, 2021.

\bibitem{qiao2021unite}
Z.~Qiao, A.~S. Christensen, M.~Welborn, F.~R. Manby, A.~Anandkumar, and T.~F.
  Miller~III, ``Unite: Unitary n-body tensor equivariant network with
  applications to quantum chemistry,'' \emph{arXiv preprint arXiv:2105.14655},
  2021.

\bibitem{brandstetter2021geometric}
J.~Brandstetter, R.~Hesselink, E.~van~der Pol, E.~Bekkers, and M.~Welling,
  ``Geometric and physical quantities improve e(3) equivariant message
  passing,'' \emph{arXiv preprint arXiv:2110.02905}, 2021.

\bibitem{schutt2021equivariant}
K.~T. Sch{\"u}tt, O.~T. Unke, and M.~Gastegger, ``Equivariant message passing
  for the prediction of tensorial properties and molecular spectra,''
  \emph{arXiv preprint arXiv:2102.03150}, 2021.

\bibitem{klicpera_gemnet_2021}
J.~Klicpera, F.~Becker, and S.~G{\"u}nnemann, ``{GemNet}: Universal directional
  graph neural networks for molecules,'' in \emph{Conference on Neural
  Information Processing Systems (NeurIPS)}, 2021.

\bibitem{ramsundar2015massively}
B.~Ramsundar, S.~Kearnes, P.~Riley, D.~Webster, D.~Konerding, and V.~Pande,
  ``Massively multitask networks for drug discovery,'' \emph{arXiv preprint
  arXiv:1502.02072}, 2015.

\bibitem{alnammi2021evaluating}
M.~Alnammi, S.~Liu, S.~S. Ericksen, G.~E. Ananiev, A.~F. Voter, S.~Guo, J.~L.
  Keck, F.~M. Hoffmann, S.~A. Wildman, and A.~Gitter, ``Evaluating scalable
  supervised learning for synthesize-on-demand chemical libraries,'' 2021.

\bibitem{jiang2021could}
D.~Jiang, Z.~Wu, C.-Y. Hsieh, G.~Chen, B.~Liao, Z.~Wang, C.~Shen, D.~Cao,
  J.~Wu, and T.~Hou, ``Could graph neural networks learn better molecular
  representation for drug discovery? a comparison study of descriptor-based and
  graph-based models,'' \emph{Journal of cheminformatics}, vol.~13, no.~1, pp.
  1--23, 2021.

\bibitem{orlando2022pyuul}
G.~Orlando, D.~Raimondi, R.~Duran-Roma{\~n}a, Y.~Moreau, J.~Schymkowitz, and
  F.~Rousseau, ``Pyuul provides an interface between biological structures and
  deep learning algorithms,'' \emph{Nature Communications}, vol.~13, no.~1, pp.
  1--9, 2022.

\bibitem{francoeur2020three}
P.~G. Francoeur, T.~Masuda, J.~Sunseri, A.~Jia, R.~B. Iovanisci, I.~Snyder, and
  D.~R. Koes, ``Three-dimensional convolutional neural networks and a
  cross-docked data set for structure-based drug design,'' \emph{Journal of
  Chemical Information and Modeling}, vol.~60, no.~9, pp. 4200--4215, 2020.

\bibitem{masci2015geodesic}
J.~Masci, D.~Boscaini, M.~Bronstein, and P.~Vandergheynst, ``Geodesic
  convolutional neural networks on riemannian manifolds,'' in \emph{Proceedings
  of the IEEE international conference on computer vision workshops}, 2015, pp.
  37--45.

\bibitem{goh2017chemception}
G.~B. Goh, C.~Siegel, A.~Vishnu, N.~O. Hodas, and N.~Baker, ``Chemception: a
  deep neural network with minimal chemistry knowledge matches the performance
  of expert-developed {QSAR/QSPR} models,'' \emph{arXiv preprint
  arXiv:1706.06689}, 2017.

\bibitem{meyer2019learning}
J.~G. Meyer, S.~Liu, I.~J. Miller, J.~J. Coon, and A.~Gitter, ``Learning drug
  functions from chemical structures with convolutional neural networks and
  random forests,'' \emph{Journal of chemical information and modeling},
  vol.~59, no.~10, pp. 4438--4449, 2019.

\bibitem{liu2018exploration}
S.~Liu, ``Exploration on deep drug discovery: Representation and learning,''
  Tech. Rep., 2018.

\bibitem{liu2019loss}
S.~Liu, Y.~Liang, and A.~Gitter, ``Loss-balanced task weighting to reduce
  negative transfer in multi-task learning,'' in \emph{Proceedings of the AAAI
  conference on artificial intelligence}, vol.~33, no.~01, 2019, pp.
  9977--9978.

\bibitem{liu2021multitask}
\BIBentryALTinterwordspacing
S.~Liu, M.~Qu, Z.~Zhang, H.~Cai, and J.~Tang, ``Multi-task learning with domain
  knowledge for molecular property prediction,'' in \emph{NeurIPS 2021 AI for
  Science Workshop}, 2021. [Online]. Available:
  \url{https://openreview.net/forum?id=6cWgY5Epwzo}
\BIBentrySTDinterwordspacing

\bibitem{liu2020structured}
S.~Liu, A.~Deac, Z.~Zhu, and J.~Tang, ``Structured multi-view representations
  for drug combinations,'' \emph{Machine Learning for Molecules Workshop at
  NeurIPS}, 2020.

\bibitem{hu2019strategies}
W.~Hu, B.~Liu, J.~Gomes, M.~Zitnik, P.~Liang, V.~Pande, and J.~Leskovec,
  ``Strategies for pre-training graph neural networks,'' in \emph{ICLR}, 2019.

\bibitem{sun2019infograph}
F.-Y. Sun, J.~Hoffmann, V.~Verma, and J.~Tang, ``{Infograph}: Unsupervised and
  semi-supervised graph-level representation learning via mutual information
  maximization,'' \emph{arXiv preprint arXiv:1908.01000}, 2019.

\bibitem{liu2021pre}
S.~Liu, H.~Wang, W.~Liu, J.~Lasenby, H.~Guo, and J.~Tang, ``Pre-training
  molecular graph representation with {3D} geometry,'' \emph{International
  Conference on Learning Representations}, 2021.

\bibitem{fu2021spear}
T.~Fu, C.~Xiao, K.~Huang, L.~M. Glass, and J.~Sun, ``{SPEAR}: self-supervised
  post-training enhancer for molecule optimization,'' in \emph{Proceedings of
  the 12th ACM Conference on Bioinformatics, Computational Biology, and Health
  Informatics}, 2021, pp. 1--10.

\bibitem{xia2022survey}
J.~Xia, Y.~Zhu, Y.~Du, and S.~Z. Li, ``A survey of pretraining on graphs:
  Taxonomy, methods, and applications,'' \emph{arXiv preprint
  arXiv:2202.07893}, 2022.

\bibitem{altae2017low}
H.~Altae-Tran, B.~Ramsundar, A.~S. Pappu, and V.~Pande, ``Low data drug
  discovery with one-shot learning,'' \emph{ACS central science}, vol.~3,
  no.~4, pp. 283--293, 2017.

\bibitem{baskin2019one}
I.~I. Baskin, ``Is one-shot learning a viable option in drug discovery?'' 2019.

\bibitem{nguyen2020meta}
C.~Q. Nguyen, C.~Kreatsoulas, and K.~M. Branson, ``Meta-learning
  initializations for low-resource drug discovery,'' \emph{arXiv preprint
  arXiv:2003.05996}, 2020.

\bibitem{rumelhart1986learning}
D.~E. Rumelhart, G.~E. Hinton, and R.~J. Williams, ``Learning representations
  by back-propagating errors,'' \emph{nature}, vol. 323, no. 6088, pp.
  533--536, 1986.

\bibitem{hochreiter1996lstm}
S.~Hochreiter and J.~Schmidhuber, ``{LSTM} can solve hard long time lag
  problems,'' \emph{Advances in neural information processing systems}, vol.~9,
  1996.

\bibitem{van2016pixel}
A.~Van~Oord, N.~Kalchbrenner, and K.~Kavukcuoglu, ``Pixel recurrent neural
  networks,'' in \emph{International conference on machine learning}.\hskip 1em
  plus 0.5em minus 0.4em\relax PMLR, 2016, pp. 1747--1756.

\bibitem{oord2016wavenet}
A.~v.~d. Oord, S.~Dieleman, H.~Zen, K.~Simonyan, O.~Vinyals, A.~Graves,
  N.~Kalchbrenner, A.~Senior, and K.~Kavukcuoglu, ``{WaveNet}: A generative
  model for raw audio,'' \emph{arXiv preprint arXiv:1609.03499}, 2016.

\bibitem{kingma2013auto}
D.~P. Kingma and M.~Welling, ``Auto-encoding variational bayes,''
  \emph{International Conference on Learning Representations}, 2013.

\bibitem{higgins2016beta}
I.~Higgins, L.~Matthey, A.~Pal, C.~Burgess, X.~Glorot, M.~Botvinick,
  S.~Mohamed, and A.~Lerchner, ``Beta-{VAE}: Learning basic visual concepts
  with a constrained variational framework,'' 2016.

\bibitem{bengio2013representation}
Y.~Bengio, A.~Courville, and P.~Vincent, ``Representation learning: A review
  and new perspectives,'' \emph{IEEE transactions on pattern analysis and
  machine intelligence}, vol.~35, no.~8, pp. 1798--1828, 2013.

\bibitem{dinh2016density}
L.~Dinh, J.~Sohl-Dickstein, and S.~Bengio, ``Density estimation using real
  {NVP},'' \emph{arXiv preprint arXiv:1605.08803}, 2016.

\bibitem{dinh2014nice}
L.~Dinh, D.~Krueger, and Y.~Bengio, ``{NICE}: Non-linear independent components
  estimation,'' \emph{ICLR (Workshop)}, 2015.

\bibitem{kingma2018glow}
D.~P. Kingma and P.~Dhariwal, ``Glow: Generative flow with invertible 1x1
  convolutions,'' \emph{Advances in Neural Information Processing Systems},
  vol.~31, pp. 10\,215--10\,224, 2018.

\bibitem{goodfellow2014generative}
I.~Goodfellow, J.~Pouget-Abadie, M.~Mirza, B.~Xu, D.~Warde-Farley, S.~Ozair,
  A.~Courville, and Y.~Bengio, ``Generative adversarial nets,'' in \emph{NIPS},
  2014, pp. 2672--2680.

\bibitem{nowozin2016f}
S.~Nowozin, B.~Cseke, and R.~Tomioka, ``f-{GAN}: Training generative neural
  samplers using variational divergence minimization,'' in \emph{Proceedings of
  the 30th International Conference on Neural Information Processing Systems},
  2016, pp. 271--279.

\bibitem{arjovsky2017wasserstein}
M.~Arjovsky, S.~Chintala, and L.~Bottou, ``Wasserstein generative adversarial
  networks,'' in \emph{International conference on machine learning}.\hskip 1em
  plus 0.5em minus 0.4em\relax PMLR, 2017, pp. 214--223.

\bibitem{karras2019style}
T.~Karras, S.~Laine, and T.~Aila, ``A style-based generator architecture for
  generative adversarial networks,'' in \emph{Proceedings of the IEEE/CVF
  conference on computer vision and pattern recognition}, 2019, pp. 4401--4410.

\bibitem{karras2020analyzing}
T.~Karras, S.~Laine, M.~Aittala, J.~Hellsten, J.~Lehtinen, and T.~Aila,
  ``Analyzing and improving the image quality of stylegan,'' in
  \emph{Proceedings of the IEEE/CVF conference on computer vision and pattern
  recognition}, 2020, pp. 8110--8119.

\bibitem{sohl2015deep}
J.~Sohl-Dickstein, E.~Weiss, N.~Maheswaranathan, and S.~Ganguli, ``Deep
  unsupervised learning using nonequilibrium thermodynamics,'' in
  \emph{International Conference on Machine Learning}.\hskip 1em plus 0.5em
  minus 0.4em\relax PMLR, 2015, pp. 2256--2265.

\bibitem{ho2020denoising}
J.~Ho, A.~Jain, and P.~Abbeel, ``Denoising diffusion probabilistic models,''
  \emph{arXiv preprint arXiv:2006.11239}, 2020.

\bibitem{weng2021diffusion}
\BIBentryALTinterwordspacing
L.~Weng, ``What are diffusion models?'' \emph{lilianweng.github.io/lil-log},
  2021. [Online]. Available:
  \url{https://lilianweng.github.io/lil-log/2021/07/11/diffusion-models.html}
\BIBentrySTDinterwordspacing

\bibitem{song2019generative}
Y.~Song and S.~Ermon, ``Generative modeling by estimating gradients of the data
  distribution,'' \emph{Advances in Neural Information Processing Systems},
  vol.~32, 2019.

\bibitem{song2020score}
Y.~Song, J.~Sohl-Dickstein, D.~P. Kingma, A.~Kumar, S.~Ermon, and B.~Poole,
  ``Score-based generative modeling through stochastic differential
  equations,'' \emph{arXiv preprint arXiv:2011.13456}, 2020.

\bibitem{hinton2002training}
G.~E. Hinton, ``Training products of experts by minimizing contrastive
  divergence,'' \emph{Neural computation}, vol.~14, no.~8, pp. 1771--1800,
  2002.

\bibitem{gutmann2010noise}
M.~Gutmann and A.~Hyv{\"a}rinen, ``Noise-contrastive estimation: A new
  estimation principle for unnormalized statistical models,'' in
  \emph{Proceedings of the Thirteenth International Conference on Artificial
  Intelligence and Statistics}, 2010, pp. 297--304.

\bibitem{song2021train}
Y.~Song and D.~P. Kingma, ``How to train your energy-based models,''
  \emph{arXiv preprint arXiv:2101.03288}, 2021.

\bibitem{zhou2019optimization}
Z.~Zhou, S.~Kearnes, L.~Li, R.~N. Zare, and P.~Riley, ``Optimization of
  molecules via deep reinforcement learning,'' \emph{Scientific reports},
  vol.~9, no.~1, pp. 1--10, 2019.

\bibitem{fu2021differentiable}
T.~Fu, W.~Gao, C.~Xiao, J.~Yasonik, C.~W. Coley, and J.~Sun, ``Differentiable
  scaffolding tree for molecular optimization,'' \emph{International Conference
  on Learning Representations}, 2022.

\bibitem{gao2020synthesizability}
W.~Gao and C.~W. Coley, ``The synthesizability of molecules proposed by
  generative models,'' \emph{Journal of chemical information and modeling},
  vol.~60, no.~12, pp. 5714--5723, 2020.

\bibitem{gao2021amortized}
W.~Gao, R.~Mercado, and C.~W. Coley, ``Amortized tree generation for bottom-up
  synthesis planning and synthesizable molecular design,'' \emph{International
  Conference on Learning Representations}, 2022.

\bibitem{brown2019guacamol}
N.~Brown, M.~Fiscato, M.~H. Segler, and A.~C. Vaucher, ``{GuacaMol}:
  benchmarking models for de novo molecular design,'' \emph{Journal of chemical
  information and modeling}, vol.~59, no.~3, pp. 1096--1108, 2019.

\bibitem{You2018-xh}
J.~You, B.~Liu, R.~Ying, V.~Pande, and J.~Leskovec, ``Graph convolutional
  policy network for goal-directed molecular graph generation,'' in
  \emph{NIPS}, 2018.

\bibitem{jin2020multi}
W.~Jin, R.~Barzilay, and T.~Jaakkola, ``Multi-objective molecule generation
  using interpretable substructures,'' in \emph{International Conference on
  Machine Learning}.\hskip 1em plus 0.5em minus 0.4em\relax PMLR, 2020, pp.
  4849--4859.

\bibitem{fu2021moler}
T.~Fu, C.~Xiao, L.~Glass, and J.~Sun, ``{MOLER}: Incorporate molecule-level
  reward to enhance deep generative model for molecule optimization,''
  \emph{IEEE Transactions on Knowledge and Data Engineering}, 2021.

\bibitem{bengio2021gflownet}
Y.~Bengio, T.~Deleu, E.~J. Hu, S.~Lahlou, M.~Tiwari, and E.~Bengio,
  ``{GFlowNet} foundations,'' \emph{CoRR}, vol. abs/2111.09266, 2021.

\bibitem{jensen2019graph}
J.~H. Jensen, ``A graph-based genetic algorithm and generative model/monte
  carlo tree search for the exploration of chemical space,'' \emph{Chemical
  science}, vol.~10, no.~12, pp. 3567--3572, 2019.

\bibitem{huang2021therapeutics}
K.~Huang, T.~Fu, W.~Gao, Y.~Zhao, Y.~Roohani, J.~Leskovec, C.~W. Coley,
  C.~Xiao, J.~Sun, and M.~Zitnik, ``Therapeutics data commons: machine learning
  datasets and tasks for therapeutics,'' \emph{NeurIPS Track Datasets and
  Benchmarks}, 2021.

\bibitem{yang2020practical}
X.~Yang, T.~K. Aasawat, and K.~Yoshizoe, ``Practical massively parallel
  monte-carlo tree search applied to molecular design,'' \emph{arXiv preprint
  arXiv:2006.10504}, 2020.

\bibitem{winter2019efficient}
R.~Winter, F.~Montanari, A.~Steffen, H.~Briem, F.~No{\'e}, and D.-A. Clevert,
  ``Efficient multi-objective molecular optimization in a continuous latent
  space,'' \emph{Chemical science}, vol.~10, no.~34, pp. 8016--8024, 2019.

\bibitem{korovina2020chembo}
K.~Korovina, S.~Xu, K.~Kandasamy, W.~Neiswanger, B.~Poczos, J.~Schneider, and
  E.~Xing, ``{ChemBO}: Bayesian optimization of small organic molecules with
  synthesizable recommendations,'' in \emph{International Conference on
  Artificial Intelligence and Statistics}.\hskip 1em plus 0.5em minus
  0.4em\relax PMLR, 2020, pp. 3393--3403.

\bibitem{notin2021improving}
P.~Notin, J.~M. Hern{\'a}ndez-Lobato, and Y.~Gal, ``Improving black-box
  optimization in vae latent space using decoder uncertainty,'' in
  \emph{Thirty-Fifth Conference on Neural Information Processing Systems},
  2021.

\bibitem{griffiths2020constrained}
R.-R. Griffiths and J.~M. Hern{\'a}ndez-Lobato, ``Constrained bayesian
  optimization for automatic chemical design using variational autoencoders,''
  \emph{Chemical science}, vol.~11, no.~2, pp. 577--586, 2020.

\bibitem{eriksson2021high}
D.~Eriksson and M.~Jankowiak, ``High-dimensional bayesian optimization with
  sparse axis-aligned subspaces,'' in \emph{Uncertainty in Artificial
  Intelligence}.\hskip 1em plus 0.5em minus 0.4em\relax PMLR, 2021, pp.
  493--503.

\bibitem{maus2022local}
N.~Maus, H.~T. Jones, J.~S. Moore, M.~J. Kusner, J.~Bradshaw, and J.~R.
  Gardner, ``Local latent space bayesian optimization over structured inputs,''
  \emph{arXiv preprint arXiv:2201.11872}, 2022.

\bibitem{jin2018junction}
W.~Jin, R.~Barzilay, and T.~Jaakkola, ``Junction tree variational autoencoder
  for molecular graph generation,'' \emph{ICML}, 2018.

\bibitem{fu2021mimosa}
T.~Fu, C.~Xiao, X.~Li, L.~M. Glass, and J.~Sun, ``{MIMOSA}: Multi-constraint
  molecule sampling for molecule optimization,'' \emph{AAAI}, 2020.

\bibitem{xie2021mars}
Y.~Xie, C.~Shi, H.~Zhou, Y.~Yang, W.~Zhang, Y.~Yu, and L.~Li, ``{MARS}: Markov
  molecular sampling for multi-objective drug discovery,'' in \emph{ICLR},
  2021.

\bibitem{dai2018syntax}
H.~Dai, Y.~Tian, B.~Dai, S.~Skiena, and L.~Song, ``Syntax-directed variational
  autoencoder for structured data,'' \emph{International Conference on Learning
  Representations ({ICLR})}, 2018.

\bibitem{kusner2017grammar}
M.~J. Kusner, B.~Paige, and J.~M. Hern{\'a}ndez-Lobato, ``Grammar variational
  autoencoder,'' in \emph{International Conference on Machine Learning}.\hskip
  1em plus 0.5em minus 0.4em\relax PMLR, 2017, pp. 1945--1954.

\bibitem{simonovsky2018graphvae}
M.~Simonovsky and N.~Komodakis, ``{GraphVAE}: Towards generation of small
  graphs using variational autoencoders,'' in \emph{International conference on
  artificial neural networks}.\hskip 1em plus 0.5em minus 0.4em\relax Springer,
  2018, pp. 412--422.

\bibitem{liu2018constrained}
Q.~Liu, M.~Allamanis, M.~Brockschmidt, and A.~Gaunt, ``Constrained graph
  variational autoencoders for molecule design,'' in \emph{NEURIPS}, 2018.

\bibitem{doi:10.1021/acs.jcim.1c01573}
\BIBentryALTinterwordspacing
A.~F. Oliveira, J.~L.~F. Da~Silva, and M.~G. Quiles, ``Molecular property
  prediction and molecular design using a supervised grammar variational
  autoencoder,'' \emph{Journal of Chemical Information and Modeling}, vol.~0,
  no.~0, p. null, 0, pMID: 35174705. [Online]. Available:
  \url{https://doi.org/10.1021/acs.jcim.1c01573}
\BIBentrySTDinterwordspacing

\bibitem{madhawa2019graphnvp}
K.~Madhawa, K.~Ishiguro, K.~Nakago, and M.~Abe, ``Graphnvp: An invertible flow
  model for generating molecular graphs,'' \emph{arXiv preprint
  arXiv:1905.11600}, 2019.

\bibitem{zang2020moflow}
C.~Zang and F.~Wang, ``{MoFlow}: an invertible flow model for generating
  molecular graphs,'' in \emph{ACM SIGKDD}, 2020, pp. 617--626.

\bibitem{shi2019graphaf}
C.~Shi, M.~Xu, Z.~Zhu, W.~Zhang, M.~Zhang, and J.~Tang, ``{GraphAF}: a
  flow-based autoregressive model for molecular graph generation,'' in
  \emph{ICLR}, 2020.

\bibitem{luo2021graphdf}
Y.~Luo, K.~Yan, and S.~Ji, ``{GraphDF}: A discrete flow model for molecular
  graph generation,'' \emph{Proceedings of the 38th International Conference on
  Machine Learning, {ICML}}, vol. 139, pp. 7192--7203, 2021.

\bibitem{bian2019deep}
Y.~Bian, J.~Wang, J.~J. Jun, and X.-Q. Xie, ``Deep convolutional generative
  adversarial network (dcgan) models for screening and design of small
  molecules targeting cannabinoid receptors,'' \emph{Molecular pharmaceutics},
  vol.~16, no.~11, pp. 4451--4460, 2019.

\bibitem{assouel2018defactor}
R.~Assouel, M.~Ahmed, M.~H. Segler, A.~Saffari, and Y.~Bengio, ``Defactor:
  Differentiable edge factorization-based probabilistic graph generation,''
  \emph{arXiv preprint arXiv:1811.09766}, 2018.

\bibitem{guimaraes2017objective}
G.~L. Guimaraes, B.~Sanchez-Lengeling, C.~Outeiral, P.~L.~C. Farias, and
  A.~Aspuru-Guzik, ``Objective-reinforced generative adversarial networks
  (organ) for sequence generation models,'' \emph{arXiv preprint
  arXiv:1705.10843}, 2017.

\bibitem{cao2018molgan}
N.~D. Cao and T.~Kipf, ``{MolGAN}: An implicit generative model for small
  molecular graphs,'' 2018.

\bibitem{popova2019molecularrnn}
M.~Popova, M.~Shvets, and J.~Oliva~et al, ``Molecularrnn: generating realistic
  molecular graphs with optimized properties,'' \emph{arXiv preprint
  arXiv:1905.13372}, 2019.

\bibitem{flam2021keeping}
D.~Flam-Shepherd, K.~Zhu, and A.~Aspuru-Guzik, ``Keeping it simple: Language
  models can learn complex molecular distributions,'' \emph{arXiv preprint
  arXiv:2112.03041}, 2021.

\bibitem{liu2021graphebm}
M.~Liu, K.~Yan, B.~Oztekin, and S.~Ji, ``Graphebm: Molecular graph generation
  with energy-based models,'' \emph{arXiv preprint arXiv:2102.00546}, 2021.

\bibitem{segler2018generating}
M.~H. Segler, T.~Kogej, C.~Tyrchan, and M.~P. Waller, ``Generating focused
  molecule libraries for drug discovery with recurrent neural networks,''
  \emph{ACS central science}, vol.~4, no.~1, pp. 120--131, 2018.

\bibitem{guo2020property}
X.~Guo, Y.~Du, and L.~Zhao, ``Property controllable variational autoencoder via
  invertible mutual dependence,'' in \emph{International Conference on Learning
  Representations}, 2020.

\bibitem{du2021deep}
Y.~Du, Y.~Wang, F.~Alam, Y.~Lu, X.~Guo, L.~Zhao, and A.~Shehu, ``Deep
  latent-variable models for controllable molecule generation,'' in \emph{2021
  IEEE International Conference on Bioinformatics and Biomedicine
  (BIBM)}.\hskip 1em plus 0.5em minus 0.4em\relax IEEE, 2021, pp. 372--375.

\bibitem{du2022interpretable}
Y.~Du, X.~Guo, A.~Shehu, and L.~Zhao, ``Interpretable molecular graph
  generation via monotonic constraints,'' \emph{SDM}, 2022.

\bibitem{kwon2021evolutionary}
Y.~Kwon, S.~Kang, Y.-S. Choi, and I.~Kim, ``Evolutionary design of molecules
  based on deep learning and a genetic algorithm,'' \emph{Scientific reports},
  vol.~11, no.~1, pp. 1--11, 2021.

\bibitem{nigam2021janus}
A.~Nigam, R.~Pollice, and A.~Aspuru-Guzik, ``{JANUS}: parallel tempered genetic
  algorithm guided by deep neural networks for inverse molecular design,''
  \emph{arXiv preprint arXiv:2106.04011}, 2021.

\bibitem{blaschke2020reinvent}
T.~Blaschke, J.~Ar{\'u}s-Pous, H.~Chen, C.~Margreitter, C.~Tyrchan,
  O.~Engkvist, K.~Papadopoulos, and A.~Patronov, ``{REINVENT} 2.0: an ai tool
  for de novo drug design,'' \emph{Journal of Chemical Information and
  Modeling}, vol.~60, no.~12, pp. 5918--5922, 2020.

\bibitem{wang2021multi}
J.~Wang, C.-Y. Hsieh, M.~Wang, X.~Wang, Z.~Wu, D.~Jiang, B.~Liao, X.~Zhang,
  B.~Yang, Q.~He \emph{et~al.}, ``Multi-constraint molecular generation based
  on conditional transformer, knowledge distillation and reinforcement
  learning,'' \emph{Nature Machine Intelligence}, vol.~3, no.~10, pp. 914--922,
  2021.

\bibitem{ahn2020guiding}
S.~Ahn, J.~Kim, H.~Lee, and J.~Shin, ``Guiding deep molecular optimization with
  genetic exploration,'' \emph{Advances in neural information processing
  systems}, vol.~33, pp. 12\,008--12\,021, 2020.

\bibitem{yang2017chemts}
X.~Yang, J.~Zhang, K.~Yoshizoe, K.~Terayama, and K.~Tsuda, ``{ChemTS}: an
  efficient python library for de novo molecular generation,'' \emph{Science
  and technology of advanced materials}, vol.~18, no.~1, pp. 972--976, 2017.

\bibitem{dollar2021attention}
O.~Dollar, N.~Joshi, D.~A. Beck, and J.~Pfaendtner, ``Attention-based
  generative models for de novo molecular design,'' \emph{Chemical Science},
  vol.~12, no.~24, pp. 8362--8372, 2021.

\bibitem{lim2020scaffold}
J.~Lim, S.-Y. Hwang, S.~Moon, S.~Kim, and W.~Y. Kim, ``Scaffold-based molecular
  design with a graph generative model,'' \emph{Chemical science}, vol.~11,
  no.~4, pp. 1153--1164, 2020.

\bibitem{pmlr-v119-jin20a}
\BIBentryALTinterwordspacing
W.~Jin, D.~Barzilay, and T.~Jaakkola, ``Hierarchical generation of molecular
  graphs using structural motifs,'' in \emph{Proceedings of the 37th
  International Conference on Machine Learning}, ser. Proceedings of Machine
  Learning Research, H.~D. III and A.~Singh, Eds., vol. 119.\hskip 1em plus
  0.5em minus 0.4em\relax PMLR, 13--18 Jul 2020, pp. 4839--4848. [Online].
  Available: \url{https://proceedings.mlr.press/v119/jin20a.html}
\BIBentrySTDinterwordspacing

\bibitem{samanta2020nevae}
B.~Samanta, A.~De, G.~Jana, V.~G{\'o}mez, P.~K. Chattaraj, N.~Ganguly, and
  M.~Gomez-Rodriguez, ``Nevae: A deep generative model for molecular graphs,''
  \emph{Journal of machine learning research. 2020 Apr; 21 (114): 1-33}, 2020.

\bibitem{chenthamarakshan2020cogmol}
V.~Chenthamarakshan, P.~Das, S.~C. Hoffman, H.~Strobelt, I.~Padhi, K.~W. Lim,
  B.~Hoover, M.~Manica, J.~Born, T.~Laino \emph{et~al.}, ``Cogmol:
  target-specific and selective drug design for covid-19 using deep generative
  models,'' \emph{arXiv preprint arXiv:2004.01215}, 2020.

\bibitem{das2021accelerated}
P.~Das, T.~Sercu, K.~Wadhawan, I.~Padhi, S.~Gehrmann, F.~Cipcigan,
  V.~Chenthamarakshan, H.~Strobelt, C.~Dos~Santos, P.-Y. Chen \emph{et~al.},
  ``Accelerated antimicrobial discovery via deep generative models and
  molecular dynamics simulations,'' \emph{Nature Biomedical Engineering},
  vol.~5, no.~6, pp. 613--623, 2021.

\bibitem{jin2018learning}
W.~Jin, K.~Yang, R.~Barzilay, and T.~Jaakkola, ``Learning multimodal
  graph-to-graph translation for molecule optimization,'' 2018.

\bibitem{fu2020core}
T.~Fu, C.~Xiao, and J.~Sun, ``{CORE}: Automatic molecule optimization using
  copy and refine strategy,'' \emph{AAAI}, 2020.

\bibitem{maziarka2020mol}
{\L}.~Maziarka, A.~Pocha, J.~Kaczmarczyk, K.~Rataj, T.~Danel, and
  M.~Warcho{\l}, ``{Mol-CycleGAN}: a generative model for molecular
  optimization,'' \emph{Journal of Cheminformatics}, vol.~12, no.~1, pp. 1--18,
  2020.

\bibitem{prykhodko2019novo}
O.~Prykhodko, S.~V. Johansson, P.-C. Kotsias, J.~Ar{\'u}s-Pous, E.~J. Bjerrum,
  O.~Engkvist, and H.~Chen, ``A de novo molecular generation method using
  latent vector based generative adversarial network,'' \emph{Journal of
  Cheminformatics}, vol.~11, no.~1, pp. 1--13, 2019.

\bibitem{yang2021hit}
S.~Yang, D.~Hwang, S.~Lee, S.~Ryu, and S.~J. Hwang, ``Hit and lead discovery
  with explorative rl and fragment-based molecule generation,'' \emph{Advances
  in Neural Information Processing Systems}, vol.~34, 2021.

\bibitem{nigam2021beyond}
A.~Nigam, R.~Pollice, M.~Krenn, G.~dos Passos~Gomes, and A.~Aspuru-Guzik,
  ``Beyond generative models: superfast traversal, optimization, novelty,
  exploration and discovery ({STONED}) algorithm for molecules using
  {SELFIES},'' \emph{Chemical science}, vol.~12, no.~20, pp. 7079--7090, 2021.

\bibitem{hoffman2020optimizing}
S.~C. Hoffman, V.~Chenthamarakshan, K.~Wadhawan, P.-Y. Chen, and P.~Das,
  ``Optimizing molecules using efficient queries from property evaluations,''
  \emph{Nature Machine Intelligence}, pp. 1--11, 2021.

\bibitem{du2022interpreting}
\BIBentryALTinterwordspacing
Y.~Du, X.~Liu, S.~Liu, and B.~Zhou, ``Interpreting molecule generative models
  for interactive molecule discovery,'' in \emph{{ELLIS} ML4Molecules
  Workshop}, 2021, under review. [Online]. Available:
  \url{https://openreview.net/forum?id=6gLEKETxUWp}
\BIBentrySTDinterwordspacing

\bibitem{mansimov2019molecular}
E.~Mansimov, O.~Mahmood, S.~Kang, and K.~Cho, ``Molecular geometry prediction
  using a deep generative graph neural network,'' \emph{Scientific reports},
  vol.~9, no.~1, pp. 1--13, 2019.

\bibitem{simm2019generative}
G.~N. Simm and J.~M. Hern{\'a}ndez-Lobato, ``A generative model for molecular
  distance geometry,'' \emph{arXiv preprint arXiv:1909.11459}, 2019.

\bibitem{xu2021end}
M.~Xu, W.~Wang, S.~Luo, C.~Shi, Y.~Bengio, R.~Gomez-Bombarelli, and J.~Tang,
  ``An end-to-end framework for molecular conformation generation via bilevel
  programming,'' 2021.

\bibitem{zhu2022direct}
J.~Zhu, Y.~Xia, C.~Liu, L.~Wu, S.~Xie, T.~Wang, Y.~Wang, W.~Zhou, T.~Qin, H.~Li
  \emph{et~al.}, ``Direct molecular conformation generation,'' \emph{arXiv
  preprint arXiv:2202.01356}, 2022.

\bibitem{shi2021learning}
C.~Shi, S.~Luo, M.~Xu, and J.~Tang, ``Learning gradient fields for molecular
  conformation generation,'' \emph{Proceedings of the 38th International
  Conference on Machine Learning, {ICML}}, vol. 139, pp. 9558--9568, 2021.

\bibitem{luo2021predicting}
S.~Luo, C.~Shi, M.~Xu, and J.~Tang, ``Predicting molecular conformation via
  dynamic graph score matching,'' \emph{Advances in Neural Information
  Processing Systems}, vol.~34, 2021.

\bibitem{xu2021geodiff}
M.~Xu, L.~Yu, Y.~Song, C.~Shi, S.~Ermon, and J.~Tang, ``{GeoDiff}: A geometric
  diffusion model for molecular conformation generation,'' in
  \emph{International Conference on Learning Representations}, 2021.

\bibitem{ganea2021geomol}
O.~Ganea, L.~Pattanaik, C.~Coley, R.~Barzilay, K.~Jensen, W.~Green, and
  T.~Jaakkola, ``{GeoMol}: Torsional geometric generation of molecular 3d
  conformer ensembles,'' \emph{Advances in Neural Information Processing
  Systems}, vol.~34, 2021.

\bibitem{chan2020bokei}
L.~Chan, G.~R. Hutchison, and G.~M. Morris, ``{BOKEI}: Bayesian optimization
  using knowledge of correlated torsions and expected improvement for conformer
  generation,'' \emph{Physical Chemistry Chemical Physics}, vol.~22, no.~9, pp.
  5211--5219, 2020.

\bibitem{gebauer2019symmetry}
N.~Gebauer, M.~Gastegger, and K.~Sch{\"u}tt, ``Symmetry-adapted generation of
  3d point sets for the targeted discovery of molecules,'' \emph{Advances in
  Neural Information Processing Systems}, vol.~32, 2019.

\bibitem{roney2021generating}
J.~P. Roney, P.~Maragakis, P.~Skopp, and D.~E. Shaw, ``Generating realistic 3d
  molecules with an equivariant conditional likelihood model,'' 2021.

\bibitem{simm2020reinforcement}
G.~Simm, R.~Pinsler, and J.~M. Hern{\'a}ndez-Lobato, ``Reinforcement learning
  for molecular design guided by quantum mechanics,'' in \emph{International
  Conference on Machine Learning}.\hskip 1em plus 0.5em minus 0.4em\relax PMLR,
  2020, pp. 8959--8969.

\bibitem{simm2020symmetry}
G.~N. Simm, R.~Pinsler, G.~Cs{\'a}nyi, and J.~M. Hern{\'a}ndez-Lobato,
  ``Symmetry-aware actor-critic for 3d molecular design,'' in
  \emph{International Conference on Learning Representations}, 2020.

\bibitem{nesterov3dmolnet}
V.~Nesterov, M.~Wieser, and V.~Roth, ``{3DMolNet}: A generative network for
  molecular structures.''

\bibitem{luo2021autoregressive}
Y.~Luo and S.~Ji, ``An autoregressive flow model for 3d molecular geometry
  generation from scratch,'' in \emph{International Conference on Learning
  Representations}, 2021.

\bibitem{durrant2013autogrow}
J.~D. Durrant, S.~Lindert, and J.~A. McCammon, ``Autogrow 3.0: an improved
  algorithm for chemically tractable, semi-automated protein inhibitor
  design,'' \emph{Journal of Molecular Graphics and Modelling}, vol.~44, pp.
  104--112, 2013.

\bibitem{masuda2020generating}
T.~Masuda, M.~Ragoza, and D.~R. Koes, ``Generating 3d molecular structures
  conditional on a receptor binding site with deep generative models,''
  \emph{arXiv preprint arXiv:2010.14442}, 2020.

\bibitem{li2021structure}
Y.~Li, J.~Pei, and L.~Lai, ``Structure-based de novo drug design using 3d deep
  generative models,'' \emph{Chemical science}, vol.~12, no.~41, pp.
  13\,664--13\,675, 2021.

\bibitem{luo20213d}
S.~Luo, J.~Guan, J.~Ma, and J.~Peng, ``A {3D} generative model for
  structure-based drug design,'' in \emph{Thirty-Fifth Conference on Neural
  Information Processing Systems}, 2021.

\bibitem{mendez2021geometric}
O.~M{\'e}ndez-Lucio, M.~Ahmad, E.~A. del Rio-Chanona, and J.~K. Wegner, ``A
  geometric deep learning approach to predict binding conformations of
  bioactive molecules,'' \emph{Nature Machine Intelligence}, vol.~3, no.~12,
  pp. 1033--1039, 2021.

\bibitem{stark2022equibind}
H.~St{\"a}rk, O.-E. Ganea, L.~Pattanaik, R.~Barzilay, and T.~Jaakkola,
  ``{EquiBind}: Geometric deep learning for drug binding structure
  prediction,'' \emph{arXiv preprint arXiv:2202.05146}, 2022.

\bibitem{chan2019bayesian}
L.~Chan, G.~R. Hutchison, and G.~M. Morris, ``Bayesian optimization for
  conformer generation,'' \emph{Journal of cheminformatics}, vol.~11, no.~1,
  pp. 1--11, 2019.

\bibitem{joshi20213d}
R.~P. Joshi, N.~W. Gebauer, M.~Bontha, M.~Khazaieli, R.~M. James, J.~B. Brown,
  and N.~Kumar, ``{3D}-scaffold: A deep learning framework to generate 3d
  coordinates of drug-like molecules with desired scaffolds,'' \emph{The
  Journal of Physical Chemistry B}, vol. 125, no.~44, pp. 12\,166--12\,176,
  2021.

\bibitem{gebauer2022inverse}
N.~W. Gebauer, M.~Gastegger, S.~S. Hessmann, K.-R. M{\"u}ller, and K.~T.
  Sch{\"u}tt, ``Inverse design of 3d molecular structures with conditional
  generative neural networks,'' \emph{Nature Communications}, vol.~13, no.~1,
  pp. 1--11, 2022.

\bibitem{wang2022generative}
W.~Wang, M.~Xu, C.~Cai, B.~K. Miller, T.~Smidt, Y.~Wang, J.~Tang, and
  R.~G{\'o}mez-Bombarelli, ``Generative coarse-graining of molecular
  conformations,'' \emph{arXiv preprint arXiv:2201.12176}, 2022.

\bibitem{zhu2022survey}
Y.~Zhu, Y.~Du, Y.~Wang, Y.~Xu, J.~Zhang, Q.~Liu, and S.~Wu, ``A survey on deep
  graph generation: Methods and applications,'' \emph{arXiv preprint
  arXiv:2203.06714}, 2022.

\bibitem{honda2019graph}
S.~Honda, H.~Akita, K.~Ishiguro, T.~Nakanishi, and K.~Oono, ``Graph residual
  flow for molecular graph generation,'' \emph{arXiv preprint
  arXiv:1909.13521}, 2019.

\bibitem{zhu2017unpaired}
J.-Y. Zhu, T.~Park, P.~Isola, and A.~A. Efros, ``Unpaired image-to-image
  translation using cycle-consistent adversarial networks,'' in
  \emph{Proceedings of the IEEE international conference on computer vision},
  2017, pp. 2223--2232.

\bibitem{du2021graphgt}
Y.~Du, S.~Wang, X.~Guo, H.~Cao, S.~Hu, J.~Jiang, A.~Varala, A.~Angirekula, and
  L.~Zhao, ``{GraphGT}: Machine learning datasets for graph generation and
  transformation,'' in \emph{Thirty-fifth Conference on Neural Information
  Processing Systems Datasets and Benchmarks Track (Round 2)}, 2021.

\bibitem{ramakrishnan2014quantum}
R.~Ramakrishnan, P.~O. Dral, M.~Rupp, and O.~A. Von~Lilienfeld, ``Quantum
  chemistry structures and properties of 134 kilo molecules,'' \emph{Scientific
  data}, vol.~1, no.~1, pp. 1--7, 2014.

\bibitem{ruddigkeit2012enumeration}
L.~Ruddigkeit, R.~Van~Deursen, L.~C. Blum, and J.-L. Reymond, ``Enumeration of
  166 billion organic small molecules in the chemical universe database
  {GDB}-17,'' \emph{Journal of chemical information and modeling}, vol.~52,
  no.~11, pp. 2864--2875, 2012.

\bibitem{sterling2015zinc}
T.~Sterling and J.~J. Irwin, ``{ZINC} 15--ligand discovery for everyone,''
  \emph{Journal of chemical information and modeling}, vol.~55, no.~11, pp.
  2324--2337, 2015.

\bibitem{polykovskiy2020molecular}
D.~Polykovskiy, A.~Zhebrak, B.~Sanchez-Lengeling, S.~Golovanov, O.~Tatanov,
  S.~Belyaev, R.~Kurbanov, A.~Artamonov, V.~Aladinskiy, M.~Veselov
  \emph{et~al.}, ``Molecular sets ({MOSES}): a benchmarking platform for
  molecular generation models,'' \emph{Frontiers in pharmacology}, 2020.

\bibitem{gaulton2012chembl}
A.~Gaulton, L.~J. Bellis, A.~P. Bento, J.~Chambers, M.~Davies, A.~Hersey,
  Y.~Light, S.~McGlinchey, D.~Michalovich, B.~Al-Lazikani \emph{et~al.},
  ``{ChEMBL}: a large-scale bioactivity database for drug discovery,''
  \emph{Nucleic acids research}, vol.~40, no.~D1, pp. D1100--D1107, 2012.

\bibitem{hachmann2011harvard}
J.~Hachmann, R.~Olivares-Amaya, S.~Atahan-Evrenk, C.~Amador-Bedolla, R.~S.
  S{\'a}nchez-Carrera, A.~Gold-Parker, L.~Vogt, A.~M. Brockway, and
  A.~Aspuru-Guzik, ``The harvard clean energy project: large-scale
  computational screening and design of organic photovoltaics on the world
  community grid,'' \emph{The Journal of Physical Chemistry Letters}, vol.~2,
  no.~17, pp. 2241--2251, 2011.

\bibitem{blum2009970}
L.~C. Blum and J.-L. Reymond, ``970 million druglike small molecules for
  virtual screening in the chemical universe database gdb-13,'' \emph{Journal
  of the American Chemical Society}, vol. 131, no.~25, pp. 8732--8733, 2009.

\bibitem{zhusurvey}
Y.~Zhu, W.~Xu, J.~Zhang, Y.~Du, J.~Zhang, Q.~Liu, C.~Yang, and S.~Wu, ``A
  survey on graph structure learning: Progress and opportunities.''

\bibitem{berman2000protein}
H.~M. Berman, J.~Westbrook, Z.~Feng, G.~Gilliland, T.~N. Bhat, H.~Weissig,
  I.~N. Shindyalov, and P.~E. Bourne, ``The protein data bank,'' \emph{Nucleic
  acids research}, vol.~28, no.~1, pp. 235--242, 2000.

\bibitem{schutt2017quantum}
K.~T. Sch{\"u}tt, F.~Arbabzadah, S.~Chmiela, K.~R. M{\"u}ller, and
  A.~Tkatchenko, ``Quantum-chemical insights from deep tensor neural
  networks,'' \emph{Nature communications}, vol.~8, no.~1, pp. 1--8, 2017.

\bibitem{xu2021molecule3d}
Z.~Xu, Y.~Luo, X.~Zhang, X.~Xu, Y.~Xie, M.~Liu, K.~Dickerson, C.~Deng,
  M.~Nakata, and S.~Ji, ``{Molecule3D}: A benchmark for predicting 3d
  geometries from molecular graphs,'' \emph{arXiv preprint arXiv:2110.01717},
  2021.

\bibitem{desaphy2015sc}
J.~Desaphy, G.~Bret, D.~Rognan, and E.~Kellenberger, ``{sc-PDB}: a 3d-database
  of ligandable binding sites—10 years on,'' \emph{Nucleic acids research},
  vol.~43, no.~D1, pp. D399--D404, 2015.

\bibitem{mysinger2012directory}
M.~M. Mysinger, M.~Carchia, J.~J. Irwin, and B.~K. Shoichet, ``Directory of
  useful decoys, enhanced (dud-e): better ligands and decoys for better
  benchmarking,'' \emph{Journal of medicinal chemistry}, vol.~55, no.~14, pp.
  6582--6594, 2012.

\bibitem{irwin2012zinc}
J.~J. Irwin, T.~Sterling, M.~M. Mysinger, E.~S. Bolstad, and R.~G. Coleman,
  ``{ZINC}: a free tool to discover chemistry for biology,'' \emph{Journal of
  chemical information and modeling}, vol.~52, no.~7, pp. 1757--1768, 2012.

\bibitem{benhenda2017chemgan}
M.~Benhenda, ``{ChemGAN} challenge for drug discovery: can ai reproduce natural
  chemical diversity?'' \emph{arXiv preprint arXiv:1708.08227}, 2017.

\bibitem{kullback1951information}
S.~Kullback and R.~A. Leibler, ``On information and sufficiency,'' \emph{The
  annals of mathematical statistics}, vol.~22, no.~1, pp. 79--86, 1951.

\bibitem{preuer2018frechet}
K.~Preuer, P.~Renz, T.~Unterthiner, S.~Hochreiter, and G.~Klambauer,
  ``Fr{\'e}chet chemnet distance: a metric for generative models for molecules
  in drug discovery,'' \emph{Journal of chemical information and modeling},
  vol.~58, no.~9, pp. 1736--1741, 2018.

\bibitem{gretton2012kernel}
A.~Gretton, K.~M. Borgwardt, M.~J. Rasch, B.~Sch{\"o}lkopf, and A.~Smola, ``A
  kernel two-sample test,'' \emph{The Journal of Machine Learning Research},
  vol.~13, no.~1, pp. 723--773, 2012.

\bibitem{landrum2013rdkit}
G.~Landrum \emph{et~al.}, ``{RDKit}: A software suite for cheminformatics,
  computational chemistry, and predictive modeling,'' 2013.

\bibitem{li2018multi}
Y.~Li, L.~Zhang, and Z.~Liu, ``Multi-objective de novo drug design with
  conditional graph generative model,'' \emph{Journal of cheminformatics},
  vol.~10, no.~1, pp. 1--24, 2018.

\bibitem{olivecrona2017molecular}
M.~Olivecrona, T.~Blaschke, O.~Engkvist, and H.~Chen, ``Molecular de-novo
  design through deep reinforcement learning,'' \emph{Journal of
  cheminformatics}, vol.~9, no.~1, p.~48, 2017.

\bibitem{trott2010autodock}
O.~Trott and A.~J. Olson, ``{AutoDock} {Vina}: improving the speed and accuracy
  of docking with a new scoring function, efficient optimization, and
  multithreading,'' \emph{Journal of computational chemistry}, vol.~31, no.~2,
  pp. 455--461, 2010.

\bibitem{guo2021deep}
X.~Guo, Y.~Du, and L.~Zhao, ``Deep generative models for spatial networks,'' in
  \emph{Proceedings of the 27th ACM SIGKDD Conference on Knowledge Discovery \&
  Data Mining}, 2021, pp. 505--515.

\bibitem{du2022disentangled}
Y.~Du, X.~Guo, H.~Cao, Y.~Ye, and L.~Zhao, ``Disentangled spatiotemporal graph
  generative models,'' \emph{AAAI}, 2022.

\bibitem{fu2022hint}
T.~Fu, K.~Huang, C.~Xiao, L.~M. Glass, and J.~Sun, ``{HINT}: Hierarchical
  interaction network for clinical-trial-outcome predictions,'' \emph{Cell
  Patterns}, p. 100445, 2022.

\bibitem{fu2021probabilistic}
T.~Fu, C.~Xiao, C.~Qian, L.~M. Glass, and J.~Sun, ``Probabilistic and dynamic
  molecule-disease interaction modeling for drug discovery,'' in
  \emph{Proceedings of the 27th ACM SIGKDD Conference on Knowledge Discovery \&
  Data Mining}, 2021, pp. 404--414.

\end{thebibliography}
